\documentclass{article}

\usepackage{bm}
\usepackage{subfiles}

\usepackage{amsmath,amsfonts,bm}

\def\eqref#1{equation~\ref{#1}}

\def\1{\bm{1}}

\def\vd{{\bm{d}}}

\def\vg{{\bm{g}}}

\def\vs{{\bm{s}}}
\def\vt{{\bm{t}}}
\def\vu{{\bm{u}}}
\def\vv{{\bm{v}}}

\def\vx{{\bm{x}}}

\def\mA{{\bm{A}}}

\def\mD{{\bm{D}}}

\def\mI{{\bm{I}}}

\def\mS{{\bm{S}}}

\def\mW{{\bm{W}}}
\def\mX{{\bm{X}}}

\DeclareMathAlphabet{\mathsfit}{\encodingdefault}{\sfdefault}{m}{sl}
\SetMathAlphabet{\mathsfit}{bold}{\encodingdefault}{\sfdefault}{bx}{n}
\newcommand{\tens}[1]{\bm{\mathsfit{#1}}}
\def\tA{{\tens{A}}}

\def\tI{{\tens{I}}}

\def\tK{{\tens{K}}}

\def\tQ{{\tens{Q}}}

\def\tV{{\tens{V}}}

\def\tX{{\tens{X}}}

\def\gD{{\mathcal{D}}}

\def\gG{{\mathcal{G}}}

\def\sA{{\mathbb{A}}}
\def\sB{{\mathbb{B}}}

\def\sD{{\mathbb{D}}}

\def\sR{{\mathbb{R}}}

\def\sV{{\mathbb{V}}}

\newcommand{\R}{\mathbb{R}}

\DeclareMathOperator*{\argmax}{arg\,max}

\usepackage{multicol}       
\usepackage{multirow}
\usepackage[table]{xcolor} 
\usepackage{subcaption}

\usepackage{microtype}
\usepackage{graphicx}
\usepackage{subcaption}
\usepackage{booktabs} 

\usepackage{hyperref}

\usepackage[accepted]{styles/icml2025}

\usepackage[ruled,vlined]{algorithm2e}
\SetKwComment{Comment}{/* }{ */}
\usepackage{amsmath}
\usepackage{amssymb}
\usepackage{wrapfig}
\usepackage{mathtools}
\usepackage{amsthm}

\usepackage[capitalize,noabbrev]{cleveref}

\theoremstyle{plain}

\theoremstyle{definition}

\theoremstyle{remark}

\usepackage[textsize=tiny]{todonotes}

\icmltitlerunning{Token Merge with Attention for Diffusion Models}

\begin{document}

\twocolumn[
\icmltitle{ToMA: Token Merge with Attention for Diffusion Models}




\icmlsetsymbol{equal}{*}

\begin{icmlauthorlist}
\icmlauthor{Wenbo Lu}{equal,yyy}
\icmlauthor{Shaoyi Zheng}{equal,yyy}
\icmlauthor{Yuxuan Xia}{yyy}
\icmlauthor{Shengjie Wang}{yyy}
\end{icmlauthorlist}

\icmlaffiliation{yyy}{Department of Computer Science, New York University}

\icmlcorrespondingauthor{Wenbo Lu}{wenbo.lu@nyu.edu}
\icmlcorrespondingauthor{Shaoyi Zheng}{sz3684@nyu.edu}
\icmlcorrespondingauthor{Yuxuan Xia}{yx2432@nyu.edu}
\icmlcorrespondingauthor{Shenjie Wang}{sw5973@nyu.edu}

\icmlkeywords{Machine Learning, ICML}

\vskip 0.3in
]



\printAffiliationsAndNotice{\icmlEqualContribution} 


\begin{abstract}
Diffusion models excel in high-fidelity image generation but face scalability limits due to transformers’ quadratic attention complexity. Plug-and-play token reduction methods like ToMeSD and ToFu reduce FLOPs by merging redundant tokens in generated images but rely on GPU-inefficient operations (e.g., sorting, scattered writes), introducing overheads that negate theoretical speedups when paired with optimized attention implementations (e.g., FlashAttention). To bridge this gap, we propose \textbf{To}ken \textbf{M}erge with \textbf{A}ttention (ToMA), an off-the-shelf method that redesigns token reduction for GPU-aligned efficiency, with three key contributions: 1) a reformulation of token merge as a submodular optimization problem to select diverse tokens; 2) merge/unmerge as an attention-like linear transformation via GPU-friendly matrix operations; and 3) exploiting latent locality and sequential redundancy (pattern reuse) to minimize overhead. ToMA reduces SDXL/Flux generation latency by 24\%/23\% (with DINO \( \Delta < \) 0.07), outperforming prior methods. This work bridges the gap between theoretical and practical efficiency for transformers in diffusion. Code available at \href{https://github.com/WenboLuu/ToMA}{\texttt{github.com/WenboLuu/ToMA}}.\looseness=-1
\end{abstract}

\section{Introduction}

Diffusion models~\cite{ho2020denoising, song2021scorebased, dhariwal2021diffusion} have revolutionized high-fidelity image generation. Yet, their reliance on transformer architectures—notably in U-ViT~\cite{bao2023all} and DiT~\cite{peebles2023scalable}—introduces a bottleneck: the quadratic complexity of self-attention scales prohibitively with token counts, exacerbating latency across denoising steps.

Efforts to accelerate transformers broadly fall into two categories: attention optimization (e.g., Flash Attention~\cite{dao2023flashattention}, xformers~\cite{xFormers2022}) and token reduction. While there is extensive research on token reduction for \emph{discriminative tasks}, including Token Merge~\cite{Bolya2023-ne}, AdaViT~\cite{yin2022vit}, and later improvements like Diffrate~\cite{chen2023diffrate}, these approaches often employ irreversible token pruning or merging, which is unsuitable for \emph{generative tasks}. 

Generative architectures like diffusion models impose stricter constraints: token counts must be restored after merging (i.e., ``unmerging'') to preserve spatial consistency for iterative refinement. Earlier attempts, including ToMeSD~\cite{Bolya-tomesd}, Token Pruning~\cite{tokenprune}, and ToFu~\cite{ToFU} adapt token reduction to this setting but suffer from a critical flaw: their merge/prune (and unmerge/fill) operations rely on GPU-inefficient primitives (e.g., sorting, scattered memory writes). This problem intensifies when paired with highly optimized attention implementations like those in the \texttt{diffusers} framework~\cite{von-platen-etal-2022-diffusers}.

In such scenarios, the attention mechanism, once the primary bottleneck, is streamlined to approach hardware efficiency limits, leaving only marginal time reductions attainable. Consequently, these potential gains are dwarfed by the computation costs of unoptimized merging logic, which now dominates the computation time.
In other words, the overhead introduced by previous token merging methods becomes negligible only when applied with fast implementations of attention, thereby preventing practical speed-ups.

\begin{figure*}[t!]
    \centering
    \includegraphics[width=0.95\linewidth]{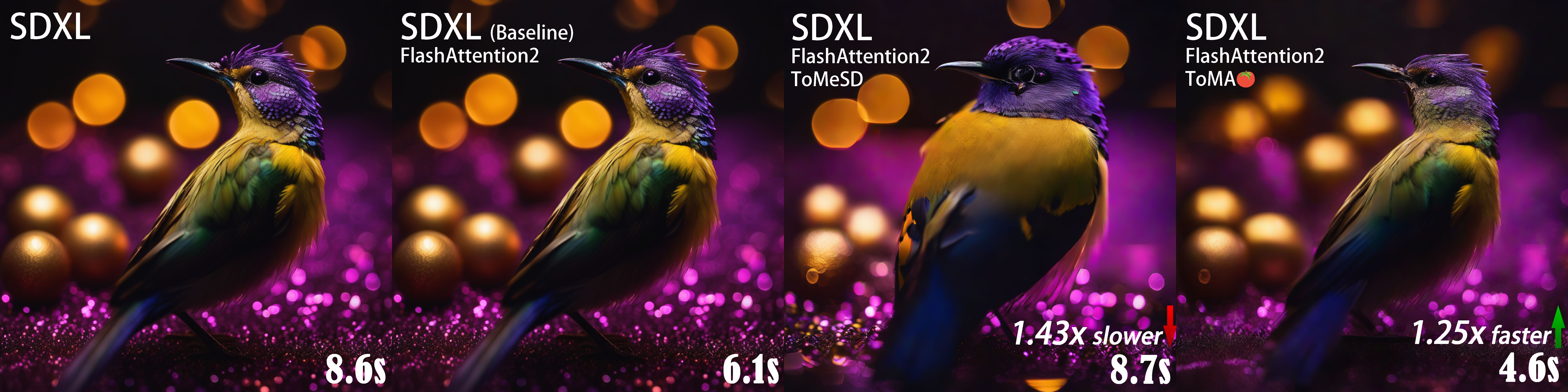}
    \caption{Comparison on \texttt{SDXL-base} with four configurations (left to right): Original, +FA2, +ToMeSD, +ToMA (on top of FA2, ratio=0.5). While ToMeSD fails to speed up due to overhead, ToMA achieves significant acceleration with negligible loss in image quality.}
    \label{fig: first illustration}
\end{figure*}

In this paper, we propose \textbf{To}ken \textbf{M}erge with \textbf{A}ttention (\textbf{ToMA}), a training-free framework that bridges this gap by rethinking token merge for GPU-aligned operation. To achieve practical speed-up without image degradation, we reformulate token merge as a submodular optimization problem, leveraging its theoretical guarantees and algorithmic toolkit. Specifically, ToMA uses our GPU-optimized facility location algorithm to select a diverse and representative set of ``destination'' tokens. Merge is then formulated as an attention-like linear transformation for efficient aggregation of non-destination tokens, with unmerge implemented via its inverse transformation, making full use of hardware-friendly matrix multiplications. This approach down-projects the latent space while preserving critical information, accelerating transformers with negligible overhead or quality loss. To further minimize computational overhead, we exploit two intrinsic properties of diffusion models:
\begin{itemize}
    \vspace{-7pt}
    \item Latent Space Locality: Tokens exhibit spatial coherence, allowing parallel merging within non-overlapping local windows (e.g., 8×8 patches).
    \item Sequential Redundancy: Merge patterns persist across 1)~adjacent denoising timesteps; and 2)~consecutive transformer layers. We amortize the overhead by reusing merge patterns over multiple steps and layers.
    \vspace{-7pt}
\end{itemize}

Theoretical guarantees from submodular optimization ensure that ToMA’s token selection approximates optimal coverage. At the same time, its co-design with GPU execution paradigms (e.g., batched matrix operations) eliminates costly operations inherent in prior methods. This synergy translates to \textbf{real-world speedup}, rather than theoretical FLOP reductions only. For example, ToMA reduces the total generation time for \texttt{SDXL-base} by 24\% and \texttt{Flux.1-dev} by 23\% with negligible degradation of image quality (change in DINO score \( < 0.07 \)), outperforming previous work like ToMeSD and ToFu, which either fails to accelerate modern attention implementations or introduce artifacts at comparable compression rates. Our contributions are summarized: \looseness=-1

\begin{itemize}
    \vspace{-7pt}
    \item Algorithmic Innovation: A submodular optimization framework for token merging, ensuring provably representative token selection to enhance quality.

    \item System Co-Design: GPU-aligned implementation strategies leveraging invertible, attention-like operations to exploit latent space locality and temporal redundancy, minimizing computational overhead.
    
    \item Empirical Validation: ToMA achieves at least \( 1.24 \times \) practical speedup when paired with FlashAttention2, State-of-the-art results across different diffusion models (e.g., \texttt{SDXL-base}, \texttt{Flux.1-dev}) and GPU architectures (NVIDIA RTX6000, V100, RTX8000).
\end{itemize}

\begin{figure*}[t!]
    \centering
    \includegraphics[width=0.95\textwidth]{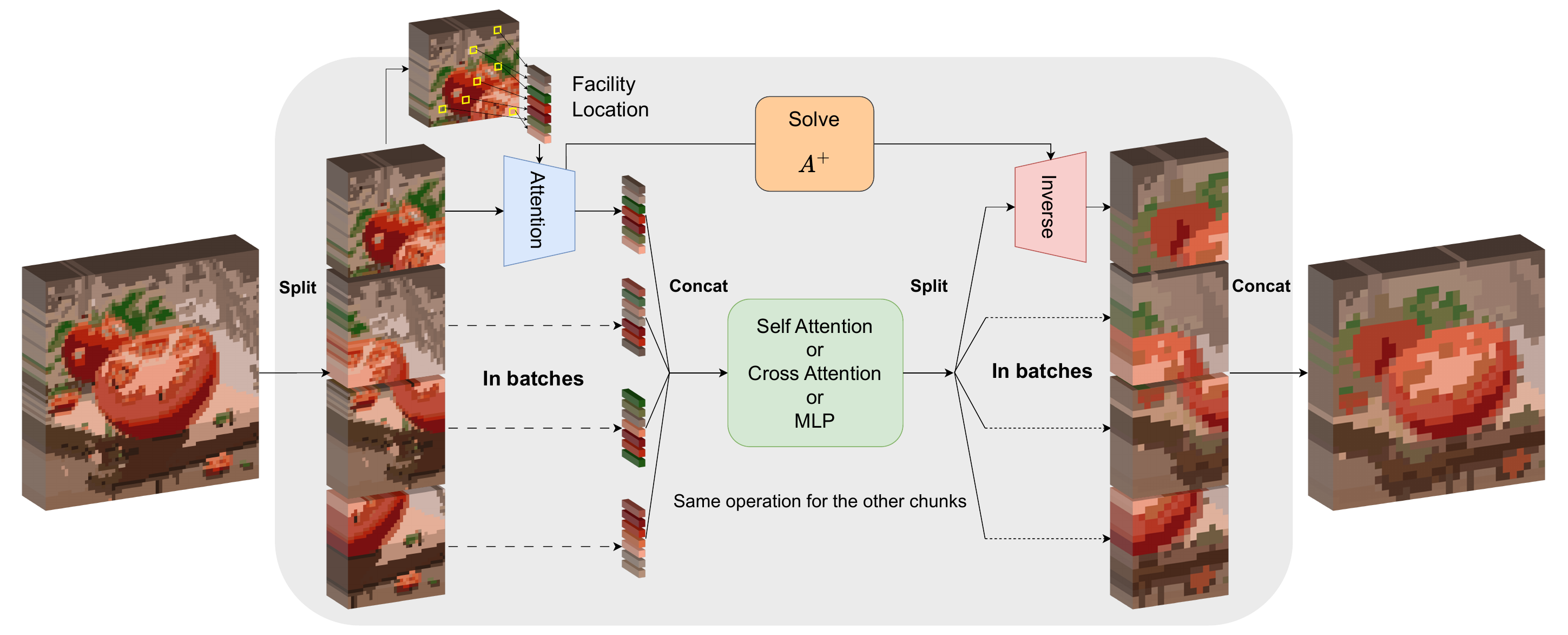}
    \caption{\small Architectural overview of ToMA. The framework consists of three key stages: (1) \textbf{Facility Location Algorithm} identifies the best representative token set $D \subset N$ through submodular optimization to maximize representational diversity; (2) \textbf{Attention (Merge)} constructs an efficient low-rank attention matrix that maps \( N\rightarrow D \) via a linear transformation for transformer computation (SelfAttn, CrossAttn, MLP) in the reduced space; (3) \textbf{Inverse (Unmerge)} applies the pseudo-inverse to recover full-resolution features \( D\rightarrow N \). The pipeline operates through localized processing of latent space regions with parallel batch optimization for efficiency.}
    \label{fig: toma_architecture}
    \vspace{-5pt}
\end{figure*}


\section{Related Work}

\paragraph{Efficient Vision Transformer}
Vision Transformers~\cite{dosovitskiy2021imageworth16x16words} face computational challenges due to quadratic attention complexity. Recent efforts to mitigate this fall into four categories: compact architectures (e.g., Swin Transformer~\cite{liu2021swin}, PVT~\cite{wang2021pyramid}), pruning strategies like X-Pruner~\cite{yu2023x}, knowledge distillation such as DeiT~\cite{touvron2021training}, and post-training quantization~\cite{liu2021posttrainingquantizationvisiontransformer}. Complementary efforts also explore combined techniques \cite{Papa_2024} to unify these paradigms. While effective, most solutions require retraining and remain inherently token-centric. In contrast, ToMA introduces training-free token merging, operating orthogonally, thus enabling seamless integration without conflict.

\paragraph{Learned Token Reduction}
Most learned token reduction involves training auxiliary models to assess the importance of tokens in the input data. DynamicViT~\cite{Rao2021-fm} employs a lightweight MLP to generate pruning masks based on input token features. A-ViT~\cite{Yin2022-hr} computes halting probabilities using specific channels of token features to determine the necessity of further processing.
Whereas these methods often require additional training for the auxiliary modules, our approach is directly applicable, offering a more generalizable solution.


\paragraph{Heuristic Token Reduction}
Heuristic token reduction strategies can be applied to ViTs without additional training. Our method also falls into this category, making the approaches below natural baselines for comparison.
Adaptive Token Sampling (ATS)~\cite{Fayyaz2022-fz} keeps tokens most similar to the class token, which limits its use in pixel-level generation tasks where a class token is absent.
Token Downsampling (ToDo)~\cite{smith2024todo} downsamples only the key–value in attention, skipping queries, which limits acceleration and causes fine-grained detail loss due to uniform spatial pooling.
Token Merging for Stable Diffusion (ToMeSD)~\cite{Bolya-tomesd} forms source–destination pairs within fixed or randomly tiled regions and greedily matches them based on similarity, performing unweighted merging followed by a simple unmerge that copies the destination embedding back to each source position.
ToFu~\cite{ToFU} builds on ToMeSD by dynamically deciding, for each layer, whether to merge or prune tokens according to a linearity test, thereby combining the benefits of both operations.

Prior token-reduction methods share two flaws: GPU-inefficient matching/merging operations and heuristic designs lacking theoretical guarantees of information preservation. By contrast, ToMA employs a GPU-friendly attention-like linear projection whose destinations are chosen via a submodular objective, providing both hardware efficiency and a principled foundation. Moreover, ToMA remains compatible with orthogonal blended schemes (alternating between pruning and merging) such as ToFu, allowing additional speed–quality trade-offs when combined.


\section{Preliminaries}

\subsection{Attention Notation}

The standard Scaled Dot-Product Attention (SDPA) mechanism is widely adopted in mainstream diffusion models. For clarity, we define the following notations: $B$ for batch size, $N$ for sequence length, $d$ for feature dimension, $D$ for the number of all destination tokens, $\tX \in \R^{B \times N \times d}$ for the attention input latent tensor, and $\tQ, \tK, \tV \in \R^{B \times N \times d}$ for query, key, and value tensors, respectively, projected from $\tX$. \looseness=-1

The SDPA operation is defined as
\begin{equation}
	\text{SDPA}(\tQ, \tK, \tV, \tau) = \text{softmax}\left(\frac{\tQ\tK^T}{\tau \sqrt{d}}\right)\tV.
    \label{eq: sdpa}
\end{equation}

\subsection{Submodularity}
\label{sec: submodularity}

A submodular function~\cite{fujishige2005submodular} is a set function \( f: 2^{\sV} \rightarrow \mathbb{R} \) defined over subsets of the ground set \( \sV \). It satisfies the diminishing returns property, which states that the marginal gain of adding a new element \( v \) to the set decreases as the context set grows. Mathematically:

For any subsets \( \sA \subseteq \sB \subseteq \sV \) and element \( v \in \sV \setminus \sB \): 
\[
	f(v | \sA) \geq f(v | \sB),
\]
where the marginal gain \( f(\cdot |\cdot) \) is defined as:
\[
	f(v | \sA) \equiv f(\sA \cup \{v\}) - f(\sA).
\]

\begin{algorithm}[tbh]
	\caption{Greedy Algorithm}
	\label{alg: greedy algorithm for facility location}
	\KwIn{Ground set \( \sV \), submodular function \( f: 2^{\sV} \to \mathbb{R} \), and budget \( k \)}
	\KwOut{Selected subset \( \sA \) of size at most \( k \)}
	    
	Initialize \( \sA \gets \emptyset \)\;
	    
	\For{\( i = 1 \) to \( k \)}{
		Select \( v^* = \arg\max_{v' \in \sV \setminus \sA} f(v' | \sA) \)\;
		Update \( \sA \gets \sA \cup \{v^*\} \)\;
	}
	    
	\Return \( \sA \)\;
\end{algorithm}

This property makes submodular functions well-suited for modeling diversity and coverage in subset selection problems. Naturally, this leads to the canonical problem of submodular maximization under a cardinality constraint:
\[
	\max_{\sA\subseteq \sV} \; f(\sA) \quad \text{s.t.} \; |\sA| \leq k.
\]

An intuitive approach is the greedy algorithm (Alg.~\ref{alg: greedy algorithm for facility location}), which guarantees a \((1 - 1/e)\)-approximation of the optimal solution~\cite{nemhauser1978analysis}. Starting with \( \sA = \emptyset \), the algorithm iteratively selects the element with the highest marginal gain until the constraint \( |\sA| = k \) is reached.

\section{Method}



Standard token merging reduces the number of tokens processed in Transformer blocks by identifying and aggregating similar tokens, thereby enabling theoretical speedups proportional to the merge ratio and the model's computational complexity (see Appendix for analysis). It works by selecting destination tokens from the full token set and merging nearby tokens into them based on similarity scores. During the unmerge step, the values of the merged tokens are redistributed to their original positions, preserving fidelity.

Our lightweight and efficient framework, ToMA, improves upon standard token merge at three key stages:  
1)~Destination Token Selection – efficiently identifying the most representative tokens to serve as merge targets;  
2)~Token Merge – performing merge operations as a linear transformation, guided by similarity scores computed via attention;  
3)~Token Unmerge – restoring merged tokens after passing through core computational modules (e.g., Attention, MLP) to original positions through reversed linear transformation.

To achieve further speedups, ToMA a)~exploits the spatial locality of the latent space to parallelize operations within local regions and, b)~shares merge-related computations across layers and iterations to reduce runtime overhead.

\subsection{Submodular-Based Destination Selection}
Let \(\mS\) be the cosine similarity matrix between all hidden states \( \mX \), where the \( \mS_{ij} \) entry represents the similarity between the \( i \)~th token and the \( j \)~th token, namely \(\mS_{ij} \equiv \cos(\mX_i, \mX_j)\).  
We denote the set of all tokens as \( \sV \) (ground set in submodular optimization) and the set of chosen destination tokens as \( \sD \).  

\begin{equation}
	f_{\text{FL}}(\sD) = \sum_{\vv_i\in \sV} \max_{\vv_j \in \sD} \mS_{ij}
	\label{eq: facility location}
\end{equation}

The submodular function used for destination token selection is the Facility Location function (FL), as shown in Eq.~\ref{eq: facility location}.  
\(f_{\text{FL}}(\cdot)\) quantifies how well a subset \(\sD\) of destination tokens represents the full token set \(\sV\) by summing, for each token \(\vv_i \in \sV\), the maximum similarity \(\mS_{ij}\) to any destination token \(\vv_j \in \sD\). 
Intuitively, this corresponds to asking: for each token in the ground set, how well does the selected subset represent it?
A higher value of \(f_{\text{FL}}(\sD)\) implies that every token in \(\sV\) is closely matched by a representative in \(\sD\), making \(\sD\) a compact and diverse summary of the input. This naturally aligns with the objective of token merging, where we aim to preserve global semantic structure using a reduced set of tokens. Notably, our framework is modular—other submodular functions can be substituted for \(f_{\text{FL}}\) to customize the selection behavior. \looseness=-1

The submodular nature of \(f_{\text{FL}}\) provides a theoretical guarantee: greedy maximization yields a near-optimal subset with provably minimal information loss, as discussed in Sec.~\ref{sec: submodularity}. When optimizing the destination set \(\sD\) using the greedy algorithm (Alg.~\ref{alg: greedy algorithm for facility location}), we iteratively select the token that provides the largest marginal gain \(f_{\text{FL}}(\vv | \sD')\) with respect to the current set \(\sD'\). This marginal gain can be efficiently computed (see Appendix~\ref{sec: efficient coverage derivation} for derivation) as:
\[
\argmax_{\vv_i \notin \sD'} \sum_{j=1}^{N} \max\big(0, \mS_{ij} - \boldsymbol{m}_j(\sD')\big),
\]
where \(\boldsymbol{m}_j(\sD') = \max_{\vv_k \in \sD'} \mS_{j,k}\) is a cached vector that stores, for each token \(\vv_j \in \sV\), the maximum similarity to any token currently in \(\sD'\). This caching enables efficient updates: after adding a new token to \(\sD'\), \(\boldsymbol{m}\) can be incrementally updated in constant time per token. 
Importantly, all these operations—computing similarities, caching, and evaluating marginal gains—can be expressed in matrix form, supported by our derivation in Appendix~\ref{sec: efficient coverage derivation}, making them highly suitable for parallel execution on GPUs. We include our efficient GPU implementation for the greedy algorithm in Appendix~\ref{sec: facility location gpu}. Also, though the iterative nature of submodular optimization is inherently unavoidable, we manage to parallelize this process by breaking it into smaller ground sets in Sec.~\ref {sec: further speedup}.  \looseness=-1

While there exist more advanced submodular maximization methods such as the Lazier-than-Lazy Greedy algorithm~\cite{mirzasoleiman2015lazier}, their reliance on operations like random subset sampling introduces irregular memory access patterns that are inefficient on GPUs. Therefore, the standard greedy approach strikes a practical balance between solution quality and hardware efficiency. 

\subsection{(Un)merge with Attention}
We begin by formulating token merge in its exact form and then show how it naturally generalizes to a linear projection over the input token space, paving the way for our proposed attention-like merging.

Let \( \mX \in \mathbb{R}^{N \times d} \) denote the input matrix of \(N\) token embeddings, each of dimension \(d\). Suppose we select a set of \(\sD\) destination tokens with indices \( \gD = \{j_1, \dots, j_D\} \subseteq \{1, \dots, N\} \), and partition the input tokens into \(D\) disjoint groups:
\[
\gG_1, \gG_2, \dots, \gG_D,
\]
\[
\text{s.t.} \quad \bigcup_{k=1}^D \gG_k = \{1, \dots, N\}, 
\quad \gG_k \cap \gG_l = \emptyset \quad \text{for } k \ne l.
\]

Each merged token \( \vx^{\text{merged}}_k \in \sR^d \) is computed by aggregating the tokens assigned to group \( \gG_k \). In the simplest case, this is done via uniform averaging:
\[
\vx^{\text{merged}}_k = \sum_{i \in \gG_k} \frac{1}{|\gG_k|} \vx_i,
\]
or more generally, we allow token-specific weights \( \alpha_{k,i} \) with normalization:
\[
\vx^{\text{merged}}_k = \sum_{i \in \gG_k} \frac{\alpha_{k,i}}{Z_k} \vx_i, \quad \text{where} \quad Z_k = \sum_{i \in \gG_k} \alpha_{k,i}.
\]

This formulation can be unified by expressing the merged tokens as a linear projection over the input matrix:
\[
\mX_{\text{merged}} = \mW \mX \in \mathbb{R}^{D \times d},
\]
where \( \mW \in \mathbb{R}^{D \times N} \) is a non-negative weight matrix with \( W_{ik} \) indicating the contribution of token \(k\) to destination \(i\). This subsumes both hard merging schemes (e.g., ToMeSD, where each row of \( \mW \) is one-hot) and soft merging (ToMA, where \( \mW \) contains normalized attention scores).

This linear formulation not only provides a principled interpretation of token merging but also enables efficient implementation and reuse of merge/unmerge operations (namely the weight matrix \(\mX\) across steps and layers.

\subsubsection{Merge}
Given the formulation above, naturally, one may think of constructing the merge weight matrix \( \mW \) via attention. Specifically, we treat the destination tokens as queries and all input tokens as keys and values, using Scaled Dot-Product Attention (SDPA) to produce similarity scores. These attention scores serve as soft merge assignments.

Let \( \mX \in \mathbb{R}^{N \times d} \) denote the input token matrix, and let \( \mD \in \mathbb{R}^{D \times d} \) be the destination token matrix, formed by selecting a subset of \(D\) token embeddings from \( \mX \). We compute an attention matrix \( \mA \in \mathbb{R}^{D \times N} \) between destinations (as queries) and all input tokens (as keys) using SDPA with a temperature parameter \( \tau \):
\[
\mA = \text{softmax}\left( \frac{\mD \mX^\top}{\tau} \right).
\]

It is important to note that softmax is applied column-wise rather than row-wise, as a token should not be decomposed into components whose aggregate exceeds 100\%. By including an extra identity matrix \( \mI \), we obtain an exact form of SDPA (Eq.~\ref{eq: sdpa}). Note that the included \( \mI \) is functionally redundant and can be omitted in implementation:
\[
    \mA = \text{SDPA} \left( \mX, \mD, \mI, \tau \right).
    \label{eq: toma sdpa}
\]

To form a proper merge weight matrix, we normalize the attention matrix row-wise:
\[
\tilde{\mA}_{ij} = \frac{\mA_{ji}}{\sum_k \mA_{jk}},
\]
and finally, the merged token representation can be obtained via matrix multiplication:
\[
\mX_{\text{merged}} = \tilde{\mA} \mX \in \mathbb{R}^{D \times d}.
\]

Intuitively, this operation softly assigns each token to a set of destination tokens based on similarity. Highly similar source tokens contribute more to the destination representations, while dissimilar ones contribute less.

By reducing merge to matrix multiplications and the SDPA kernel, ToMA scales up with token count easily and effectively receives a free ride from ongoing GPU architecture and ML system improvements for attention, which are increasingly efficient in both computation and memory usage. \looseness=-1

\subsubsection{Unmerge}
Feeding merged tokens into core computation modules,
\[
\mX' = \textsc{Attention/MLP}(\mX) \in \mathbb{R}^{D \times d},
\]
the output we get is still \( D \), and we restore the original token resolution by applying an approximate inverse of the merge projection. Let \( \mX'  \) denote the output of core computational modules, with merged tokens as input. The goal is to reconstruct the full-resolution token matrix \( \mX'_{\text{unmerged}} \in \mathbb{R}^{N \times d} \).

A principled approach is to apply the Moore–Penrose pseudo-inverse:
\[
\mX'_{\text{unmerged}} = \tilde{\mA}^+ \mX' = \tilde{\mA}^\top(\tilde{\mA} \tilde{\mA}^\top)^{-1} \mX'.
\]
This provides a least-squares reconstruction that minimizes the error between the original and reconstructed tokens, assuming the merge–transform–unmerge process remains approximately linear. However, computing the pseudo-inverse is computationally expensive, requiring matrix decompositions such as SVD or QR.

Fortunately, under certain conditions, a much simpler and more efficient alternative is available:
\[
\mX'_{\text{unmerged}} = \tilde{\mA}^\top \mX',
\]
which happens whenever \( \tilde{\mA}\tilde{\mA}^{\top} = \mI_D \), making the pseudo-inverse \( \tilde{\mA}^{+}= \tilde{\mA}^{\top}(\tilde{\mA}\tilde{\mA}^{\top})^{-1} \) collapse to the simple transpose \( \tilde{\mA}^{\top} \). This is equivalent to saying that the rows of \( \tilde{\mA} \) are \text{orthonormal}. In ToMA, this condition is approached in practice. The facility-location selection promotes destination tokens that cover largely disjoint subsets of the source tokens, making rows of \( \tilde{\mA} \) distinct. Moreover, the low attention temperature \( \tau \) adopted sharpens the softmax distribution, concentrating each row’s mass on a few source tokens and bringing its \(\ell_2\)-norm close to one. 

Together, these properties imply:
\[
\tilde{\mA} \tilde{\mA}^\top = \mI_D + \varepsilon, \qquad \| \varepsilon \| \ll 1,
\]
so that
\[
(\tilde{\mA} \tilde{\mA}^\top)^{-1} \approx \mI_D - \varepsilon, \quad \text{and thus} \quad \tilde{\mA}^+ \approx \tilde{\mA}^\top.
\]

Empirically, \( \tilde{\mA}^\top \) remains competitive against the exact pseudo-inverse. Given this high fidelity and the substantial computational and memory savings, ToMA adopts the transpose-based unmerge \( \tilde{\mA}^\top \mX' \) as the default method.

\subsection{Further Speedups}
\label{sec: further speedup}
The overhead of ToMA arises from three main sources: 1) selecting destination tokens \( \mD \) through submodular optimization; 2) computing the attention-based (un)merge weight matrices \( \tilde{\mA} \) and \( \tilde{\mA}^+ \); and 3) applying the merge \( \tilde{\mA}\mX  \) before, and unmerge \( \tilde{\mA}^+\mX'  \) after each module inside the transformer blocks. To further reduce these overheads, we exploit the locality of the feature space, enabling these computations to be performed within localized regions. Additionally, we reduce the frequency of steps 1) and 2) by reusing destination selections and (un)merge matrices across multiple iterations and transformer layers. \looseness=-1

\subsubsection{Locality-Aware Token Merging}
\begin{figure}[t]
	\centering
	\includegraphics[width=0.5\textwidth, trim={30 50 30 110}, clip]{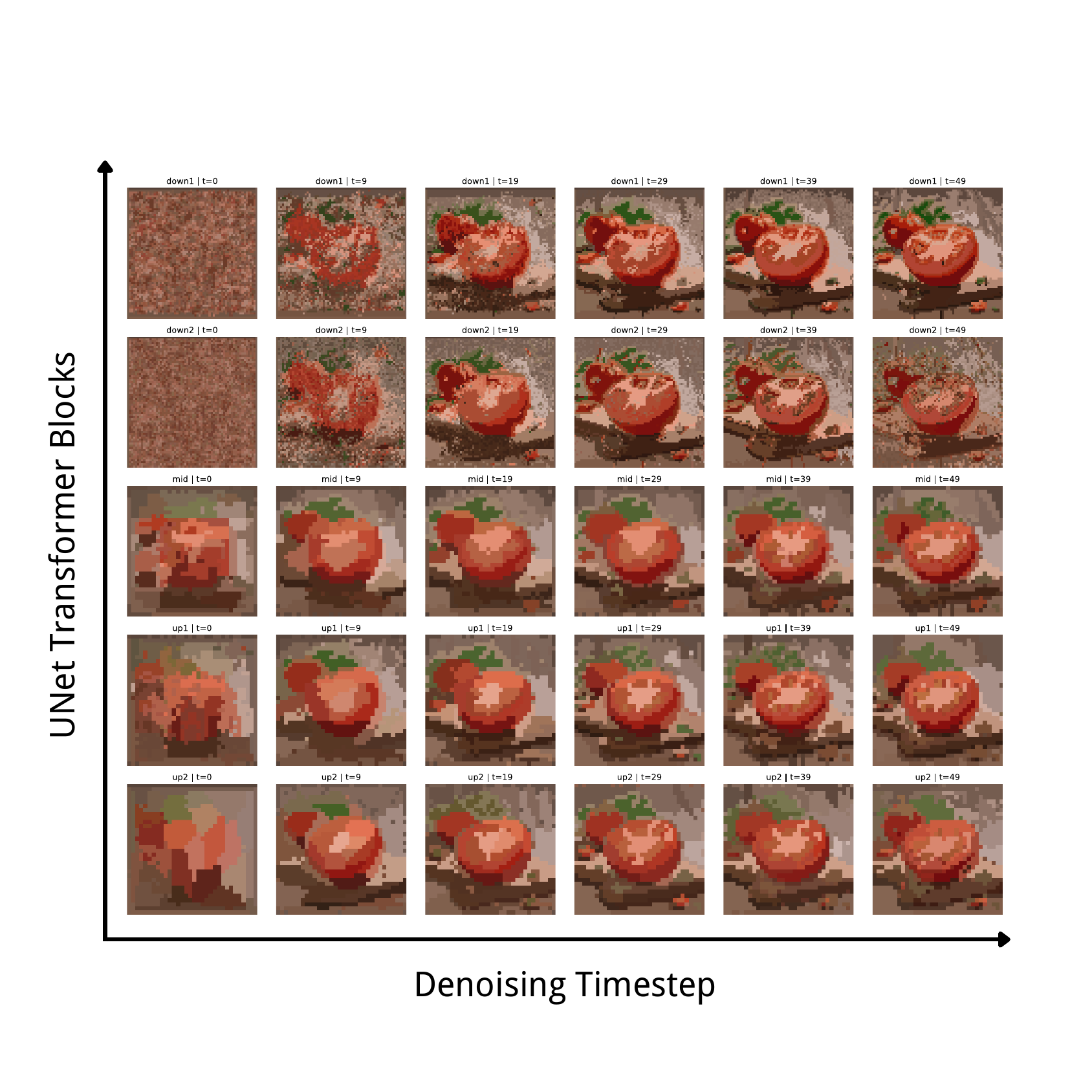}
	\caption{Re-colored \emph{k}-means clusters of U-ViT hidden states across transformer blocks and denoising timesteps. A similar visualization on DiT is provided in the Appendix~\ref{appendix: dit locality}.}
	\label{fig:unet locality}
\end{figure}

A key oversight in prior work is the \emph{locality structure} of the latent space. As shown in Fig.~\ref{fig:unet locality}, where we visualize \emph{k}-means clusters of U-ViT hidden states during the generation of a ``tomato'', the recolored tokens form a rough preview of the generated image. For example, in the early denoising steps, the clusters appear as coarse, blocky color regions that gradually refine into a recognizable tomato.

Because natural images exhibit strong local coherence--each pixel tends to resemble its immediate neighbors--their latents inherit this property. As a result, merging within a small spatial window aggregates highly similar information, preserving global structure while discarding redundancy only, and is therefore as effective as global merging. In terms of selecting destination tokens, restricting the facility location search to local regions ensures diversity within each tile and avoids competition across tiles for the same destination.

The primary benefit of this locality constraint is computational. Splitting the sequence into \(k\) equal-sized tiles yields a dominant \(1/k\) speed-up for destination selection and an even greater \(1/k^2\) reduction in computing the attention weight matrices as well as applying (un)merge. Detailed complexity analysis is provided in Appendix~\ref{sec: complexity}.

To leverage these benefits, ToMA limits destination selection and (un)merge operations to within localized regions using the two partitioning strategies below.

\paragraph{Tile-shaped Regions}  
Tokens are divided into 2-D tiles, preserving both horizontal and vertical proximity.  This layout aligns closely with image geometry and gives the best quality, albeit at a reshuffling cost on GPUs.

\paragraph{Stripe-shaped Regions}  
Tokens are grouped by rows directly, maintaining memory contiguity and enabling fast reshaping.  Although this ignores vertical proximity, it provides the highest speedup.

Both variants substantially reduce computation by operating on smaller subsets in parallel. Tile-shaped regions offer higher fidelity, while stripe-shaped regions run much faster. Despite substantial acceleration, further acceleration is possible by implementing custom tiled/stripe attention kernels, in which memory copying overhead incurred in either reshape or read as strided no longer exists. However, as the post-literature doesn't include such a low-level implementation, we decide to leave it as future work for the fairness of comparison. The full locality-aware ToMA algorithm is given in Appendix Alg.~\ref{alg: toma} in detail. \looseness=-1

\subsubsection{Reusing Destinations and Merge Weights}
\begin{figure}[t]
	\centering
	\includegraphics[width=0.95\linewidth]{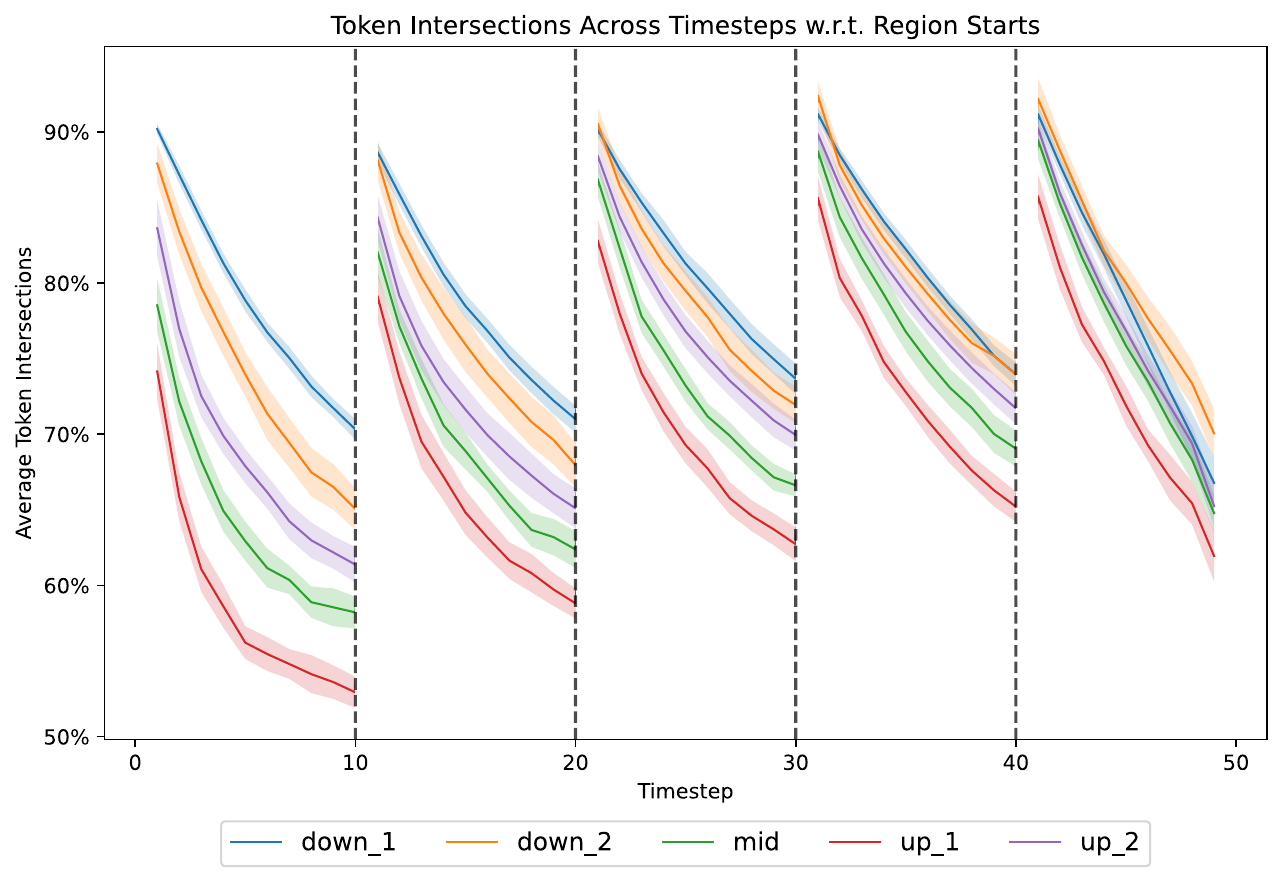}
        \vspace{-5pt}
	\caption{Average percentage of shared destination tokens at each denoising timestep relative to the first step of its 10-step interval. Each curve represents a different layer in \texttt{SDXL-base} U-ViT model, showing high overlap and gradual divergence over time.}
	\label{fig: weight sharing}
\end{figure}

Hidden states in diffusion models change gradually, so the destination tokens chosen at one step are often similar to those chosen nearby in time. Fig.~\ref{fig: weight sharing} quantifies this effect: across a 10-step window, more than half of the destinations are reused. Exploiting this redundancy, ToMA reuses the same destination set for several consecutive steps and, because merge and unmerge are linear, also reuses the associated weight matrices across layers. Sharing both the destination selection and the (un)merge matrices sharply reduces the frequency of introducing expensive similarity computation overhead while preserving image quality. \looseness=-1


\section{Experiments}
\subsection{Descriptions}

\paragraph{Setup} We evaluate ToMA on two of the most widely used diffusion models: the UNet-based \texttt{SDXL-base} and the DiT-based \texttt{Flux.1-dev}, to generate $1024 \times 1024$ images, using the \texttt{Diffusers} framework. Importantly, ToMA is architecture-agnostic and can be readily extended to other diffusion models (e.g., \texttt{SD2, SD3.5}). Prompts are drawn from the GEMRec dataset~\cite{GEMRec} and ImageNet--1K names of the classes~\cite{imagenet1k}.

\paragraph{Metrics}To assess image quality, we use CLIP-T, DINO, and FID~\cite{CLIP, dino, fid}. CLIP-T measures semantic alignment between images and prompts via cosine similarity between image-text embeddings. DINO measures perceptual consistency by comparing visual features between the original generated image and its counterpart produced with the merge method applied. FID (Fréchet Inception Distance) quantifies distributional similarity to real images based on Inception-V3 feature statistics. For FID CLIP-T and DINO in the main experiments shown in this section, we generate 3,000 images and compute scores against ground-truths from ImageNet--1K. For the ablation experiments listed in Appendix~\ref{Appendix: Ablation}, we generate images from 50 prompts with three random seeds each and report the average. FID is omitted on GEMRec due to the lack of paired images. Inference latency is reported as the median wall-clock time over 100 runs. \looseness=-1

\paragraph{Baselines} 
We compare ToMA with three heuristic token-reduction baselines: Token Merging for Stable Diffusion (\texttt{ToMe}), Token Downsampling (\texttt{ToDo}), and Token-Fusion (\texttt{ToFu}). These approaches were originally developed for UNet-based architectures and perform well in that setting. As a result, they are not compatible with DiT models, since they lack mechanisms to handle positional embeddings in DiTs. Due to this limitation, we evaluate all three baselines only on \texttt{SDXL-base}, leaving \texttt{Flux.1-dev} benchmarked solely with ToMA. In terms of implementation, we utilize the official codebases for \texttt{ToDo} and \texttt{ToMe}, and reimplement \texttt{ToFu} based on its paper, as the public code is not available. Specially, we use a fixed 75\% merge ratio for \texttt{ToDo}, corresponding to a 4-to-1 token downsampling scheme, which represents the lowest merge ratio supported by its implementation. \looseness=-1

We additionally report \textsc{TLB} (Theoretical Lower Bound), which approximates the maximum attainable speedup by reducing the number of tokens without incurring extra runtime overhead (e.g., gather tokens). To simulate this bound, we do a dummy merge--directly drop tokens and proceed with the next module by duplicating retained token features to preserve input shape, thereby isolating the theoretical benefits of token reduction while minimizing implementation-specific costs. Quality metrics are omitted for \textsc{TLB}, as cloned tokens do not yield valid outputs for evaluation. \looseness=-1

\paragraph{ToMA Variants} 
To analyze the impact of locality on destination selection and (un)merge operations, we evaluate four configurations:  
1)~\texttt{ToMA}, our default setting, which uses tile-based destination selection and global attention-based merge;  
2)~\texttt{ToMA\textsubscript{stripe}}, which restricts both destination and merge operations to within stripe regions;  
3)~\texttt{ToMA\textsubscript{tile}}, which uses tile regions for both destination and merge; and  
4)~\texttt{ToMA\textsubscript{once}}, which improves efficiency by performing (un)merge operations only once per Transformer block—at the beginning and end—rather than around each core computation module. As for the hyperparameter, stripe- and tile-based configurations use 64 stripes or tiles, respectively. We reuse the destination for 10 denoising steps and reuse merge weights for 5 steps, with each block of a given type sharing one set. No reuse across denoising timesteps in \texttt{Flux.1-dev} but within blocks of the same kind.

\subsection{Results}

\begin{table}[ht!]
	\centering
	\small
	\setlength{\tabcolsep}{3pt} 
	\resizebox{\columnwidth}{!}{ 
		\begin{tabular}{@{}c|l|rcc|ccc}
			\toprule
			\multirow{2}{*}{\textbf{Ratio}} & \multirow{2}{*}{\quad\textbf{Method}} & \multicolumn{3}{c|}{\textbf{Metrics}} & \multicolumn{3}{c}{\textbf{Sec/img \( \downarrow \)}} \\ 
			\cmidrule(lr){3-5} \cmidrule(lr){6-8}
			                  &                                     & \textbf{FID\( \downarrow \)} & \textbf{CLIP-T\( \uparrow \)} & \textbf{DINO\( \downarrow \)} & \textbf{RTX6000}         & \textbf{V100}             & \textbf{RTX8000}          \\ 
			\midrule
			\textbf{Baseline} & \texttt{SDXL}                  & 25.27                        & 29.89                        & 0                         & 6.1                      & 14.5                      & 16.1                      \\ 
			\midrule
			\multirow{5}{*}{0.25} 
			                  & \texttt{ToMA}                       & 25.72                        & 29.86                         & 0.048                         & 6.0                      & 14.3                      & 15.9                      \\
			                  & \texttt{ToMA\textsubscript{stripe}} & \cellcolor[gray]{0.9}25.17   & \cellcolor[gray]{0.9}29.90    & 0.054                         & 5.6                      & 12.6                      & 14.5                      \\
			                  & \texttt{ToMA\textsubscript{tile}}   & 25.43                        & 29.86                         & \cellcolor[gray]{0.9}0.045    & 6.2                      & 13.6                      & 15.7                      \\
			                  & \texttt{ToMA\textsubscript{once}}   & 26.31                        & 29.70                         & 0.052                         & \cellcolor[gray]{0.9}5.5 & \cellcolor[gray]{0.9}12.3 & \cellcolor[gray]{0.9}13.5 \\
			\cmidrule{2-8}
			                  & \textsc{TLB}                                 & \multicolumn{3}{c|}{--\quad\quad--\quad\quad--\quad\quad--\quad\quad--}                            & 5.2                      & 12.1                      & 9.2                       \\
			\midrule
			\multirow{5}{*}{0.50} 
			                  & \texttt{ToMA}                       & \cellcolor[gray]{0.9}28.88   & \cellcolor[gray]{0.9}29.64    & 0.068                         & 5.0                      & 11.0                      & 12.8                      \\
			                  & \texttt{ToMA\textsubscript{stripe}} & 29.11                        & 29.52                         & 0.074                         & \cellcolor[gray]{0.9}4.6 & 10.1                      & 12.0                      \\
			                  & \texttt{ToMA\textsubscript{tile}}   & 29.19                        & 29.63                         & \cellcolor[gray]{0.9}0.063    & 6.3                      & 11.1                      & 13.2                      \\
			                  & \texttt{ToMA\textsubscript{once}}   & 38.14                        & 29.06                         & 0.080                         & 4.9                      & \cellcolor[gray]{0.9}9.7  & \cellcolor[gray]{0.9}11.5 \\
			\cmidrule{2-8}
			                  & \textsc{TLB}                                 & \multicolumn{3}{c|}{--\quad\quad--\quad\quad--\quad\quad--\quad\quad--}                            & 4.0                      & 9.9                       & 7.8                       \\
			\midrule
			\multirow{5}{*}{0.75} 
			                  & \texttt{ToMA}                       & \cellcolor[gray]{0.9}58.59   & 27.96                         & 0.098                         & \cellcolor[gray]{0.9}4.3 & 8.5                       & 9.8                       \\
			                  & \texttt{ToMA\textsubscript{stripe}} & 89.93                        & 26.97                         & 0.110                         & 4.5                      & 8.0  & 9.5  \\
			                  & \texttt{ToMA\textsubscript{tile}}   & 58.90                        & \cellcolor[gray]{0.9}28.17    & \cellcolor[gray]{0.9}0.091    & 6.2                      & 9.1                       & 10.7                      \\
			                  & \texttt{ToMA\textsubscript{once}}   & 123.37                       & 24.96                         & 0.106                         & 4.9                      & \cellcolor[gray]{0.9}7.6  & \cellcolor[gray]{0.9}8.9  \\
			\cmidrule{2-8}
			                  & \textsc{TLB}                                 & \multicolumn{3}{c|}{--\quad\quad--\quad\quad--\quad\quad--\quad\quad--}                          & 3.1                      & 7.8                       & 6.5                       \\
			\bottomrule
		\end{tabular}
	}
	\caption{Performance comparison between ToMA variants and \texttt{SDXL-base} (Baseline) for generating 1024×1024 images over 50 sampling steps. Best values are highlighted, except for \textsc{TLB}. ($\uparrow$: higher better, $\downarrow$: lower better).}
	\label{tab:sdxl toma variants}
\end{table}

\paragraph{UNet Results} Table~\ref{tab:sdxl toma variants} shows that \texttt{ToMA\textsubscript{tile}} consistently performs the best on DINO score, indicating strong perceptual alignment, but is slowed down by low-level memory copying overhead during token tiling. Our manual inspection of generated images also confirms that tile-based merging yields the highest visual quality. In contrast, \texttt{ToMA\textsubscript{stripe}} benefits from faster runtimes due to its sequential memory access pattern, which enables direct reshaping without copying; however, the absence of a strong locality leads to slightly degraded image quality. \texttt{ToMA} finds a favorable trade-off, achieving up to \textbf{24\%} speedup and consistent performance across different GPU architectures. Based on this balance between efficiency and quality, we adopt it as our default method. Finally, the experimental variant \texttt{ToMA\textsubscript{once}} offers the highest acceleration by treating the entire Transformer block as a single merge unit, significantly reducing overhead. However, it produces the worst quality due to insufficient spatial-context mixing across layers. Interestingly, in some scenarios, its runtime is even lower than that of the \textsc{TLB}, likely due to less frequent memory copying compared to our dummy merge implementation.

\begin{table}[ht]
    \centering
    \small
    \setlength{\tabcolsep}{3pt} 
    \renewcommand{\arraystretch}{1.2} 
    \resizebox{\columnwidth}{!}{ 
        \begin{tabular}{@{}c|l|rcc|cr|cr@{}}
            \toprule
            \multirow{2}{*}{\textbf{Ratio}} & \multirow{2}{*}{\quad\textbf{Method}} & \multicolumn{3}{c|}{\textbf{Metrics}} & \multicolumn{2}{c|}{\textbf{RTX8000}} & \multicolumn{2}{c}{\textbf{RTX6000}} \\ 
            \cmidrule(lr){3-5} \cmidrule(lr){6-7} \cmidrule(lr){8-9}
                                           &                                      & \textbf{FID\( \downarrow \)} & \textbf{CLIP-T\( \uparrow \)} & \textbf{DINO\( \downarrow \)} & \textbf{Sec/img} & \textbf{↓$\Delta \quad$} & \textbf{Sec/img} & \textbf{↓$\Delta \quad$} \\ 
            \midrule
            \textbf{Baseline} & \texttt{Flux.1-dev}           & 31.56 & 29.03 & 0      & 59.20 & 0\%     & 21.03 & 0\%     \\ 
            \midrule
            \multirow{2}{*}{0.25} 
            & \texttt{ToMA}                 & \cellcolor[gray]{0.9}30.80 & \cellcolor[gray]{0.9}29.07 & 0.043 & \cellcolor[gray]{0.9}56.70 & \cellcolor[gray]{0.9}--4.2\% & \cellcolor[gray]{0.9}20.14 & \cellcolor[gray]{0.9}--4.2\% \\
            & \texttt{ToMA\textsubscript{tile}} & 31.49 & 29.05 & \cellcolor[gray]{0.9}0.021 & 57.47 & --2.9\% & 20.78 & --1.2\% \\
            \midrule
            \multirow{2}{*}{0.50} 
            & \texttt{ToMA}                 & \cellcolor[gray]{0.9}31.70 & 29.09 & 0.051 & \cellcolor[gray]{0.9}51.44 & \cellcolor[gray]{0.9}--13.1\% & \cellcolor[gray]{0.9}18.58 & \cellcolor[gray]{0.9}--11.6\% \\
            & \texttt{ToMA\textsubscript{tile}} & 32.95 & \cellcolor[gray]{0.9}29.19 & \cellcolor[gray]{0.9}0.032 & 53.61 & --9.4\%  & 19.61 & --6.8\%  \\
            \midrule
            \multirow{2}{*}{0.75} 
            & \texttt{ToMA}                 &\cellcolor[gray]{0.9}33.39 & 28.98 & 0.064 & \cellcolor[gray]{0.9}49.83 & \cellcolor[gray]{0.9}--15.9\% & \cellcolor[gray]{0.9}16.12 & \cellcolor[gray]{0.9}--23.4\% \\
            & \texttt{ToMA\textsubscript{tile}} & 33.88 & \cellcolor[gray]{0.9}29.34 & \cellcolor[gray]{0.9}0.045 & 49.86 & --15.8\% & 18.30 & --12.9\% \\
            \bottomrule
        \end{tabular}
    }
    \caption{Performance comparison between ToMA variants and \texttt{Flux.1-dev} (Baseline) for 1024$\times$1024 image generation (35 sampling steps). Best values are highlighted, and relative speed improvements (\(\Delta\)) are shown as \%. Negative \(\Delta\) values indicate faster inference compared to the baseline (lower is better).}
    \label{tab:flux metrics}
\end{table}

\paragraph{DiT Results}  
For the \texttt{Flux} model, we include only \texttt{ToMA} and \texttt{ToMA\textsubscript{tile}}, as stripe-based merging is incompatible with the rotary positional embedding (RoPE) used in Flux. We also exclude results on the V100 GPU due to out-of-memory (OOM) failures. As shown in Tab.~\ref{tab:flux metrics}, both ToMA variants consistently accelerate generation across GPUs, with \texttt{ToMA} achieving up to a 23.4\% speedup without compromising image quality when merging down to 50\% of tokens. Although \texttt{ToMA\textsubscript{tile}} is marginally slower due to memory overhead, it consistently offers better fidelity, evidenced by stronger CLIP-T and DINO scores. This efficiency--quality tradeoff supports our choice of \texttt{ToMA} as the default, while also highlighting the need for optimized low--level implementations to fully realize the benefits of locality-aware merging.

\begin{figure*}[ht!]
    \centering
    \begin{minipage}{0.55\linewidth}
        \centering
        \includegraphics[width=\linewidth]{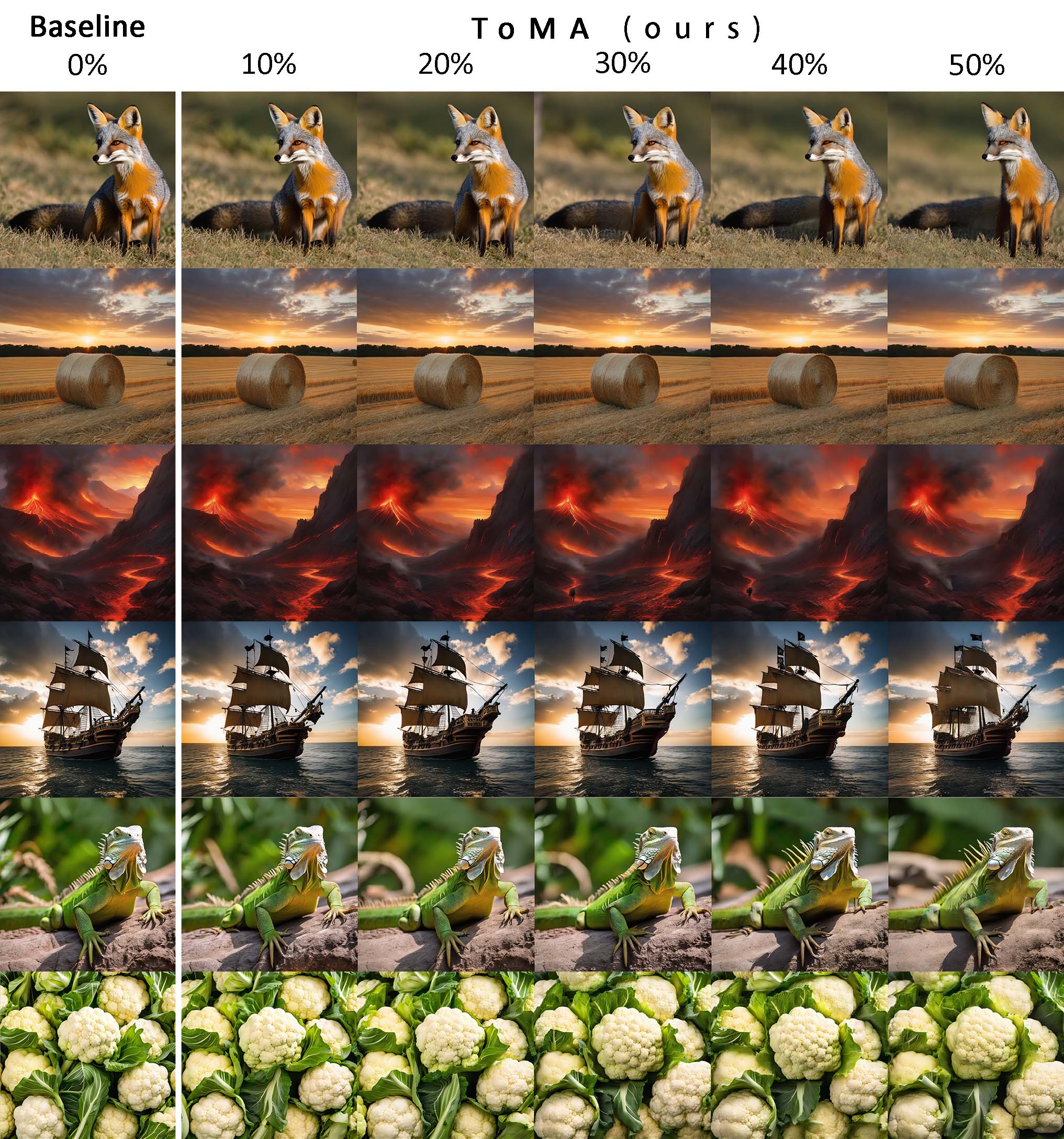}
        \subcaption*{\texttt{SDXL-base} vs. \texttt{ToMA}}
    \end{minipage}
    \hspace{0.3cm}
    \begin{minipage}{0.42\linewidth}
        \centering
        \includegraphics[width=0.9\linewidth]{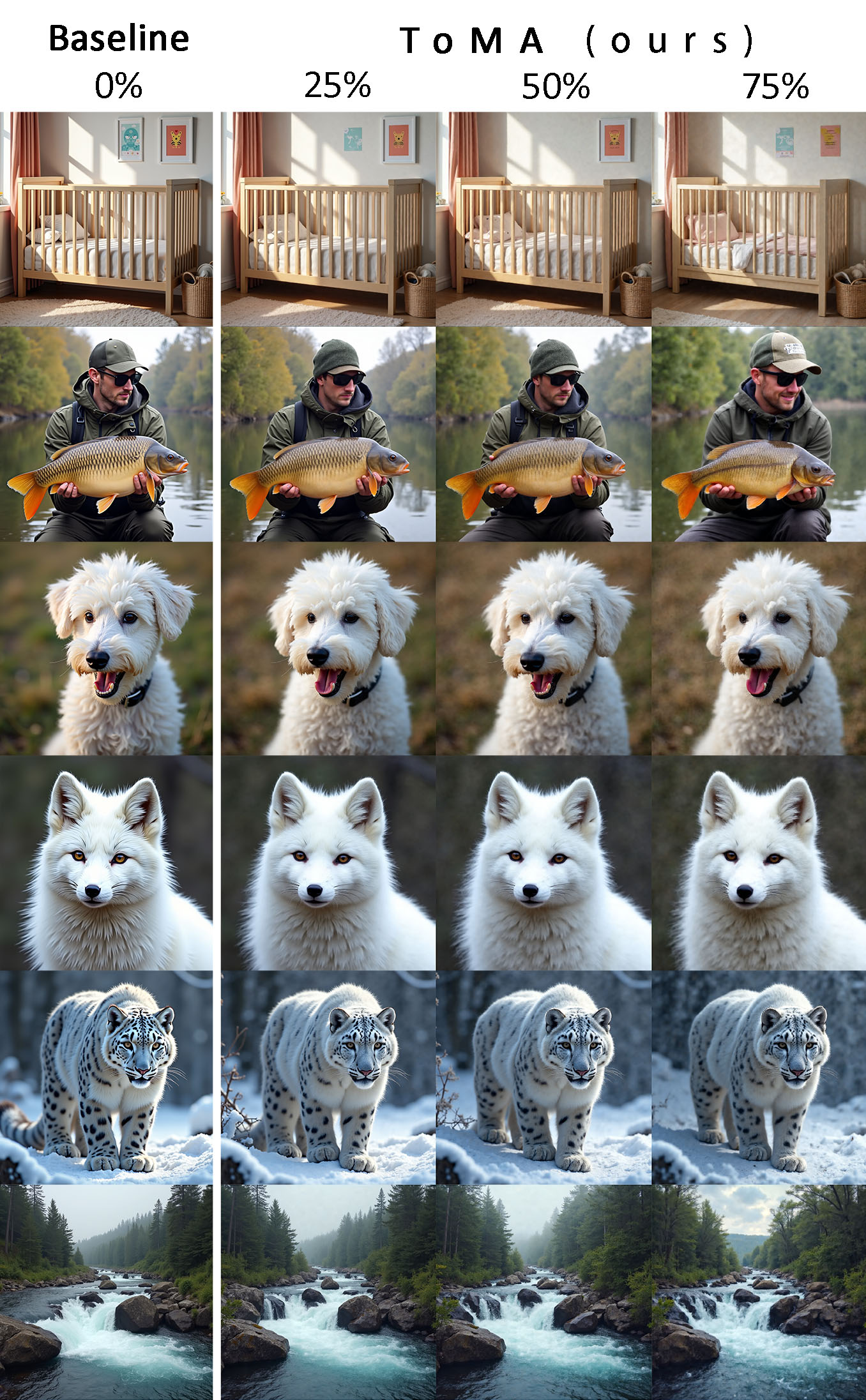}
        \subcaption*{\texttt{Flux.1-dev} vs. \texttt{ToMA}}
    \end{minipage}
    \caption{Visual comparison of baselines (\texttt{SDXL} \& \texttt{Flux}) versus (\texttt{ToMA}) under different higher token-merge ratios. Despite merging up to 50\% of tokens, \texttt{ToMA} preserves sharp details and overall scene coherence. Additional qualitative examples are provided in Appendix~\ref{Appendix: More Qualitative}.}
    \label{fig: qual result}
\end{figure*}

\begin{table}[b!]
    \centering
    \small
    \renewcommand{\arraystretch}{1} 
    \resizebox{\columnwidth}{!}{ 
    \begin{tabular}{@{}c|c|rcc|cr@{}}
        \toprule
        \textbf{Ratio} & \textbf{Method} & \textbf{FID$\downarrow$} & \textbf{CLIP-T$\uparrow$} & \textbf{DINO$\downarrow$} & \textbf{Sec/img$\downarrow$} & \textbf{↓$\Delta \quad$} \\ 
        \midrule
        \textbf{Baseline} & \texttt{SDXL}  & 25.3 & 29.89 & 0     & 6.07 & 0\% \\ 
        \midrule
        \multirow{3}{*}{0.25} 
        & \underline{\texttt{ToMA}}    & 25.7 & \cellcolor[gray]{0.9}29.86 & \cellcolor[gray]{0.9}0.048 & \cellcolor[gray]{0.9}6.03 & \cellcolor[gray]{0.9}--0.7\% \\
        & \texttt{ToMe}             & \cellcolor[gray]{0.9}25.6 & 29.86 & 0.054 & 8.66 & +42.7\% \\
        & \texttt{ToFu}             & 35.2 & 29.34 & 0.072 & 6.92 & +14.0\% \\
        \midrule
        \multirow{3}{*}{0.50} 
        & \underline{\texttt{ToMA}}    & 28.9 & 29.64 & \cellcolor[gray]{0.9}0.068 & \cellcolor[gray]{0.9}5.04 & \cellcolor[gray]{0.9}--17.0\% \\
        & \texttt{ToMe}             & \cellcolor[gray]{0.9}26.7 & \cellcolor[gray]{0.9}29.71 & 0.071 & 8.73 & +43.8\% \\
        & \texttt{ToFu}             & 142.1 & 25.04 & 0.134 & 6.83 & +12.5\% \\
        \midrule
        \multirow{4}{*}{0.75} 
        & \underline{\texttt{ToMA}}    & 58.6 & 27.96 & 0.098 & \cellcolor[gray]{0.9}4.34 & \cellcolor[gray]{0.9}--28.5\% \\
        & \texttt{ToMe}             & \cellcolor[gray]{0.9}41.2 & \cellcolor[gray]{0.9}29.09 & \cellcolor[gray]{0.9}0.084 & 8.16 & +34.4\% \\
        & \texttt{ToFu}             & 161.5 & 24.13 & 0.148 & 6.76 & +11.4\% \\
        & \texttt{ToDo}             & 68.6 & 27.60 & 0.093 & 5.67 & --6.6\% \\
        \bottomrule
    \end{tabular}
    }
    \caption{Performance comparison of ToMA, \texttt{SDXL-base} (Baseline), and other token reduction methods for generating 1024$\times$1024 images (50 sampling steps). Best values are highlighted, and relative speed improvements (\(\Delta\)) are shown as \%. (\(-\Delta\) faster, \(+\Delta\) slower; $\uparrow$: higher better, $\downarrow$: lower better).}
    \label{tab:sdxl benchmark}
\end{table}

\paragraph{Benchmark}
Table~\ref{tab:sdxl benchmark} presents a comparative evaluation of \texttt{ToMA} (our default method), alongside other token reduction strategies on the \texttt{SDXL} model. Among all methods, \texttt{ToMA} achieves the fastest generation time—up to 28.5\% improvement—while maintaining competitive perceptual quality, making it the most balanced choice. Although ToMe (Token Merge) delivers slightly better image quality, as reflected in CLIP-T and DINO scores, it suffers from severe latency due to its complex token selection mechanism, which relies on GPU-inefficient operations (e.g., sort).  ToFu (Token Fusion), while faster than ToMe with lower overhead thanks to its blended strategy that incorporates token pruning, exhibits unstable visual quality—some generations appear acceptable, but others are heavily degraded, especially at higher reduction ratios. The high FID values confirm this observation. Lastly, ToDo (Token Downsampling) offers moderate gains in both speed and fidelity but demonstrates noticeable distributional shifts, as indicated by its elevated FID score. Hwang et al. discuss the possible cause that the noise reduction due to downsampling creates a difference in the signal-to-noise ratio that leads to suboptimal performance if applied directly \yrcite{hwang2024upsample}. Overall, these results highlight \texttt{ToMA} as the most robust and well-rounded token merging strategy for efficient high-resolution image generation.

\paragraph{Qualitative Result}  
Figure~\ref{fig: qual result} provides a side‐by‐side visual comparison of images generated by the original models versus our ToMA‐accelerated variants at several token‐merge ratios. On the left, six different prompts are shown for \texttt{SDXL} and ToMA at 10\% -- 50\% token reduction. Even as the merge ratio increases, ToMA’s outputs (second through sixth columns) remain nearly indistinguishable from the originals. Similarly, on the right, we present six different prompts for \texttt{Flux} vs. ToMA at 25\%, 50\%, and 75\% token reduction. Even at 75\% merging, ToMA’s results (fourth column) faithfully reproduce key details.

\paragraph{Ablation \& Otheres.}
ToMA’s robustness is further validated by comprehensive ablations (Appendix~\ref{Appendix: Ablation}), which investigate the impact of merge frequency, tile/stripe granularity, unmerge strategies (e.g., transpose vs. pseudo-inverse), and sharing schedules. These studies confirm our design choices, such as selecting tile-based merge with 256 tiles and transpose-based unmerge for optimal speed–quality balance. Appendix~\ref{sec: memory} provides memory profiling across multiple models and sparsity levels, showing that ToMA variants incur negligible memory overhead compared to dense baselines. Appendix~\ref{sec: flop} complements this by providing a detailed FLOP analysis, saving computations up to $3.4\times$.

\section{Conclusions}
In this work, we introduce Token Merge with Attention for Diffusion Models (ToMA), a co-designed merging framework that advances prior methods in three key aspects: 1) more representative token selection via submodular optimization with theoretical guarantee; 2) a flexible and efficient merge–unmerge mechanism implemented through attention-based operations; and 3) the incorporation of locality-aware and shared merging strategies to maximize runtime gains. Together, these improvements yield substantial speedups in practice while preserving high image fidelity across different GPU and model architectures (U-ViT \& DiT), establishing ToMA as a robust and deployable solution for efficient high-resolution generation. \looseness=-1

\clearpage

\section*{Acknowledgements}
We gratefully acknowledge the New York University High Performance Computing (NYU HPC) facility for providing the GPU clusters and technical support that enabled the experiments in this work.

\section*{Impact Statement}
Our work on Token Merge with Attention (ToMA) improves the efficiency of diffusion models for image generation. While this advancement democratizes access to high-quality AI art creation, it's important to acknowledge that such technologies can be misused to generate misleading content or deepfakes. Additionally, as these models are trained on large internet-scraped datasets, there's a risk of perpetuating societal biases in the training data. We recognize these ethical considerations and emphasize the importance of responsible development and use of such technologies.



\bibliography{main}
\bibliographystyle{icml2025}

\clearpage
\appendix
\onecolumn

\section{Facility Location for Selecting Destination Tokens}
\subsection{Mathematical Foundations for Efficient Greedy Maximization}
\label{sec: efficient coverage derivation}
We begin by defining the notations used in the greedy selection procedure. Let \( \sV \) denote the full set of tokens, and let \( \sA \subseteq \sV \) be the current set of selected representative tokens. At each step, the greedy algorithm selects the next token \( \vv^* \in \sV \setminus \sA \) that maximizes the marginal gain:
\[
\vv^* = \argmax_{\vv \in (\sV \setminus \sA)} f(\vv | \sA),
\]
where the gain function \( f(\vv | \sA) \) is defined as the increase in a coverage objective when \( \vv \) is added to \( \sA \). Formally,
\begin{align*}
f(\vv | \sA)
  &= f(\sA') - f(\sA) \\
  &= \sum_{\vu \in \sV} \max_{\vu_a \in \sA'} \text{sim}(\vu, \vu_a) - \sum_{\vu \in \sV} \max_{\vu_b \in \sA} \text{sim}(\vu, \vu_b), \quad \text{where } \sA' = \{\vv\} \cup \sA.
\end{align*}

Here, $\text{sim}(\cdot, \cdot)$ denotes the similarity between two tokens (we use cosine similarity in practice for computational efficiency). The first term in the equation measures for each token $\vv \in \sV$, how well it is represented in the updated set $\sA' = \{\vv\} \cup \sA$. The second term asks the same question but with the current set $\sA$. Their difference quantifies the marginal gain of including $\vv$ in the representative set.

We can simplify the first term in the gain function by observing that:
\begin{align*}
\sum_{\vu \in \sV} \max_{\vu_a \in \sA'} \text{sim}(\vu, \vu_a)
  &= \sum_{\vu \in \sV} \max \left\{ \max_{\vu_a \in \sA} \text{sim}(\vu, \vu_a), \text{sim}(\vu, \vv) \right\},
\end{align*}
in which the identity holds because for each \( \vu \in \sV \), the maximum similarity with the updated set \( \sA' \) is either its similarity with \( \vv \), or its previous maximum over \( \sA \). So, we take the maximum of the two.

Substituting this into the definition of \( f(\vv | \sA) \), we obtain:
\begin{align*}
f(\vv | \sA)
  &= \sum_{\vu \in \sV} \left( \max \left\{ \max_{\vu_a \in \sA} \text{sim}(\vu, \vu_a), \text{sim}(\vu, \vv) \right\} - \max_{\vu_b \in \sA} \text{sim}(\vu, \vu_b) \right) \\
  &= \sum_{\vu \in \sV} \max \left\{ 0, \text{sim}(\vu, \vv) - \max_{\vu_b \in \sA} \text{sim}(\vu, \vu_b) \right\}.
\end{align*}

This simplification shows that the marginal gain of adding \( \vv \) depends only on the tokens \( \vu \in \sV \) for which \( \vv \) provides a new maximum similarity beyond what is already achieved by the current set \( \sA \).

As a result, the greedy selection objective becomes:
\[
\vv^* = \argmax_{\vv \in (\sV \setminus \sA)} \sum_{\vu \in \sV} \max \left\{ 0, \text{sim}(\vu, \vv) - \max_{\vu_b \in \sA} \text{sim}(\vu, \vu_b) \right\}.
\]

This formulation enables efficient computation of the marginal gain for each candidate token and facilitates the selection of the token that most improves the representational coverage of the current set.

\clearpage

\subsection{Facility Location Algorithm}
\label{sec: facility location gpu}
This algorithm implements a greedy approach to select $D$ tokens based on the Facility Location objective. It begins by computing the sum of similarities for each token. The first token chosen is the one with the highest sum of similarities. Then, it iteratively selects the remaining $D-1$ tokens. 

In each iteration, the algorithm computes its gain for every unselected token, representing the overall similarity improvement. The token with the highest gain is chosen and added to the selection. After each selection, the algorithm updates the maximum similarities achieved so far. This process continues until $D$ tokens are selected. 

By selecting tokens that maximize the marginal gain in similarity at each step, this approach effectively covers the input space while avoiding redundancy, ensuring a diverse and representative set of tokens.

\begin{center}
	\begin{algorithm}[h]
		\label{alg: facility location}
		\normalsize
		\SetAlgoLined
		\KwIn{Similarity matrix $\mS \in \R^{N \times N}$, number of tokens to select $D$}
		\KwOut{Selected destination token indices $\vd$}
		
		\textbf{Initialize}: $\vd \leftarrow \{\}$\;
		Sum over each row: $\vs = \sum_{j=1}^{N} \mS_{ij}$\;
		Select the first token index: $\vt_1 \leftarrow \argmax_i \vs_i$\;
		Add the greedy choice to destination tokens: $\vd \leftarrow \vd \cup \{\vt_1\}$\;
		  Create the cache vector $\boldsymbol{m}_j(\sD')\big)$ from the corresponding row in the $\mS$: $\boldsymbol{m}_j(\sD')\big) \leftarrow \mS_{\vt_1}$\;
		Set to zero to avoid re-selection: $\mS_{\vt_1} \leftarrow 0$\;
		
		\For{$k = 2$ \KwTo $D$}{
			\For{each token index $i$ not in $\vd$}{
				Compute the marginal gain efficiently with cache: $\vg = \sum_{j=1}^{N} \max\left\{0, \mS_{ij} - \boldsymbol{m}_j(\sD') \right\}$\;
			}
			Select the next token greedily: $\vt_k \leftarrow \argmax_{i \notin \vd} \vg_i$\;
			Add the newly selected token index: $\vd \leftarrow \vd \cup \{\vt_k\}$\;
			Update largest row: $\boldsymbol{m}_j(\sD') \leftarrow \max\{\boldsymbol{m}_j(\sD'), \mS_{\vt_k}\}$\;
			Set to zero to avoid re-selectoin: $\mS_{\vt_k} \leftarrow 0$\;
		}
		
		\Return $\vd$\
		
		\caption{Greedy Algorithm for Token Selection}
	\end{algorithm}
\end{center}

\clearpage

\section{Overall Detailed Algorithm of \texttt{ToMA}}
\begin{algorithm}[h]
    \caption{\texttt{ToMA} with Local Regions}
    \label{alg: toma}
    \normalsize
    \KwIn{Tensor $\tX \in \R^{B \times N \times d}$ (input sequence), $D$ (number of destination tokens),
           $\tau$ (attention temperature), $F(\cdot)$ (core computational module (e.g., MLP, Attention)}
    
    \textbf{Split into local regions}\\
    \Indp
    Partition the sequence dimension into $P$ blocks,  
    $\tX \gets (\tX_1,\ldots,\tX_P)$ with $\tX_p \in \R^{B \times N_{\text{loc}}\times d}$ and
    $N_{\text{loc}}P=N$\;
    $D_{\text{loc}} \gets D / P$\;
    $\tX \gets \tX.\mathrm{reshape}(B\!\cdot\!P,\,N_{\text{loc}},\,d)$\;
    \Indm
    
    \textbf{Step 1: Facility–location token selection}\\
    \Indp
    $(T_1,\ldots,T_{B\cdot P}) \gets \mathrm{Greedy}(f_{\text{FL}},\,D_{\text{loc}},\,\tX)$\;
    $\tX_T \gets (\tX_{1,T_1},\ldots,\tX_{B\cdot P,T_{B\cdot P}}) 
    \in \R^{B\!\cdot\!P \times D_{\text{loc}}\times d}$\;
    \Indm
    
    \textbf{Step 2: Merge}\\
    \Indp
    $\tA \gets \mathrm{SDPA}(\tX_T,\,\tX,\,\tI,\,\tau)
          \in \R^{B\!\cdot\!P \times D_{\text{loc}}\times N_{\text{loc}}}$\;
    $\tilde{\tA} \gets \tA \,/\, \tA\!\sum_{-1}$\;
    $\tX_{\mathrm{merged}} \gets \tilde{\tA}\,\tX
          \in \R^{B\!\cdot\!P \times D_{\text{loc}}\times d}$\;
    \Indm
    
    \textbf{Computational layer}\\
    \Indp
    $\tX' \gets F\!\bigl(\tX_{\mathrm{merged}}.\mathrm{reshape}(B,\,D,\,d)\bigr)$\;
    \Indm
    
    \textbf{Step 3: Unmerge}\\
    \Indp
    $\tX'_{\mathrm{unmerged}} \gets \tilde{\tA}^{\!\top}\,\tX'$\;
    Reassemble $\tX'_{\mathrm{unmerged}}$ to reverse the local-region split\;
    \Indm
    
    \Return $\tX'_{\mathrm{unmerged}}$
\end{algorithm}

\paragraph{Description.}
Algorithm~\ref{alg: toma} details a single \texttt{ToMA} layer equipped with
local-region processing.  Given an input tensor
$\tX\!\in\!\R^{B\times N\times d}$, we first shard the sequence into
$P$ equally sized local regions of length $N_{\text{loc}}$
(lines~1–3), then assign a budget of $D_{\text{loc}}$ \emph{destination
tokens} to each region (line~2).

In \textbf{Step 1} we invoke a GPU-friendly greedy
facility–location algorithm to pick, for every mini-batch $\times$ region
pair, the set of tokens whose neighbourhoods best cover the local region
(lines 5–6).  These destinations are gathered into $\tX_T$.

\textbf{Step 2} forms a
scaled-dot-product attention map $\tA$ from each destination to \emph{all}
tokens in its region, row-normalises it to
$\tilde{\tA}$, and left-multiplies $\tX$ to obtain the
merged representation $\tX_{\mathrm{merged}}$ whose sequence length is reduced
from $N$ to $D$ (lines 8–10).  Any computational block
$F(\cdot)$—e.g., \ a transformer layer or UNet block—can now process the shorter
sequence (line 12), yielding a computed reduction factor of
$N/D$ without altering the block’s internal parameters.

Now, in \textbf{Step 3}, we distribute the updated destination embeddings back to
their original token positions via $\tilde{\tA}^{\!\top}$ and undoes the
initial region partitioning (lines 14–15), so that the next layer receives a
full-resolution tensor.  Because the (un)merge operations are purely
linear and share weights across the batch, their overhead is negligible, and
they can be fused with existing attention kernels.  Repeating this procedure
layer-by-layer yields significant wall-clock speed-ups while exhibiting
scarcely any perceptible degradation in image quality.

\newpage
\section{Computational–complexity analysis}
\label{sec: complexity}
We retain explicit constant factors in the flop count because they translate directly to empirical speedups on modern GPUs.  
Throughout, \(N\) is the original sequence length, \(D\) the length \emph{after} token merging, \(d\) the embedding dimension, and  
\[
r \;=\; \frac{D}{N}\qquad(\text{merge ratio, i.e.\ the fraction of tokens \emph{kept}}).
\]

\paragraph{Baseline self–attention block.}  
Treating each matrix multiplication as a collection of dot products, the total number of scalar multiplications for a standard self–attention block is  
\[
C_{\text{base}}
  \;=\;
  4\,d^{2} N
  \;+\;
  2\,d\,N^{2}.
\]
The first term (\(Q,K,V\) projections and the output projection) scales linearly in \(N\); the second term (\(QK^\top\) and attention‐value product) scales quadratically.

\paragraph{Token–merged self–attention.}  
After reducing the token count from \(N\) to \(D\!=\!rN\), the attention cost becomes  
\[
C_{\text{attn}}(D)
  \;=\;
  4\,d^{2}D
  \;+\;
  2\,d\,D^{2}
  \;=\;
  4\,d^{2} rN
  \;+\;
  2\,d\,r^{2} N^{2}.
\]
Hence the \emph{ideal} speedup (ignoring merge overhead) is
\[
\text{Speedup}_{\text{ideal}}
  \;=\;
  \frac{C_{\text{base}}}{C_{\text{attn}}(D)}
  \;=\;
  \frac{4d + 2N}{4d\,r + 2N\,r^{2}}.
\]

\paragraph{Overheads introduced by ToMA.}  
Token merging incurs several additional costs:

\begin{itemize}
  \item \textbf{Submodular–selection overhead} (computing marginal gains for all pairs):
        \(\;C_{\text{sub}} = N^{2}d\).
  \item \textbf{Merge–attention projection} (computing pairwise weights):
        \(\;C_{\text{proj}} = N D d\).
  \item \textbf{Merge operation} (applying the computed weights to produce merged tokens):
        \(\;C_{\text{merge}} = N D d\).
  \item \textbf{Unmerge operation} with a transpose‐style redistribution:
        \(\;C_{\text{unmerge}} = N D d\).
\end{itemize}

Summing the three linear–in-\(D\) overhead terms yields
\(
C_{\text{lin}} = 3 N D d
              = 3 r N^{2} d.
\)

\paragraph{Total cost with ToMA.}  
The overall computational cost after merging is therefore
\[
C_{\text{total}}(r)
  \;=\;
  \underbrace{4\,d^{2} rN + 2\,d\,r^{2} N^{2}}_{\text{attention block}}
  \;+\;
  \underbrace{N^{2}d}_{\text{submodular}}
  \;+\;
  \underbrace{3 r N^{2} d}_{\text{merge / unmerge}}.
\]

\paragraph{Realistic speedup.}  
The practical speedup of ToMA relative to the baseline is
\[
\text{Speedup}_{\text{practical}}
  \;=\;
  \frac{C_{\text{base}}}{C_{\text{total}}(r)}
  \;=\;
  \frac{4dN + 2N^{2}}
       {4drN + N^{2}\!\bigl(1 + 3r + 2r^{2}\bigr)}.
\]

\paragraph{Discussion.}  
For practical transformer settings we have \(N \gg d\), so the quadratic terms
\(2d\,r^{2}N^{2}\) (remaining attention cost) and \(N^{2}d\) (one–shot
sub-modular selection) dominate.  Consequently, the empirical speedup is well
approximated by
\(
\text{Speedup}\;\approx\;
\bigl(2 + 4d/N\bigr)\!/\!\bigl(2r^{2} + 1 + 3r\bigr),
\)
which approaches the analytic bound
\(
\tfrac{2}{2r^{2}+1}
\)
whenever \(r \lesssim 0.5\) and \(d/N\) is small.  
At moderate merge ratios (\(r\!\in\![0.25,0.5]\)) this yields the
\(24\!-\!28\%\) latency drop we observe on SDXL/Flux.  
If we push \(r\) below \(\sim0.1\), the linear–in–\(r\) merge overhead
\(3rN^{2}d\) and the fixed
\(N^{2}d\) selection cost start to dominate, so further merging brings
diminishing returns—underscoring why our locality-aware pattern reuse
amortises these costs across multiple layers.

\newpage

\section{More Qualitative Results}
\label{Appendix: More Qualitative}
\subsection{Images Generated with SDXL}
\begin{figure}[h!tbp]
	\centering
	\begin{minipage}[t]{0.49\textwidth}
		\centering
		\includegraphics[width=0.85\textwidth]{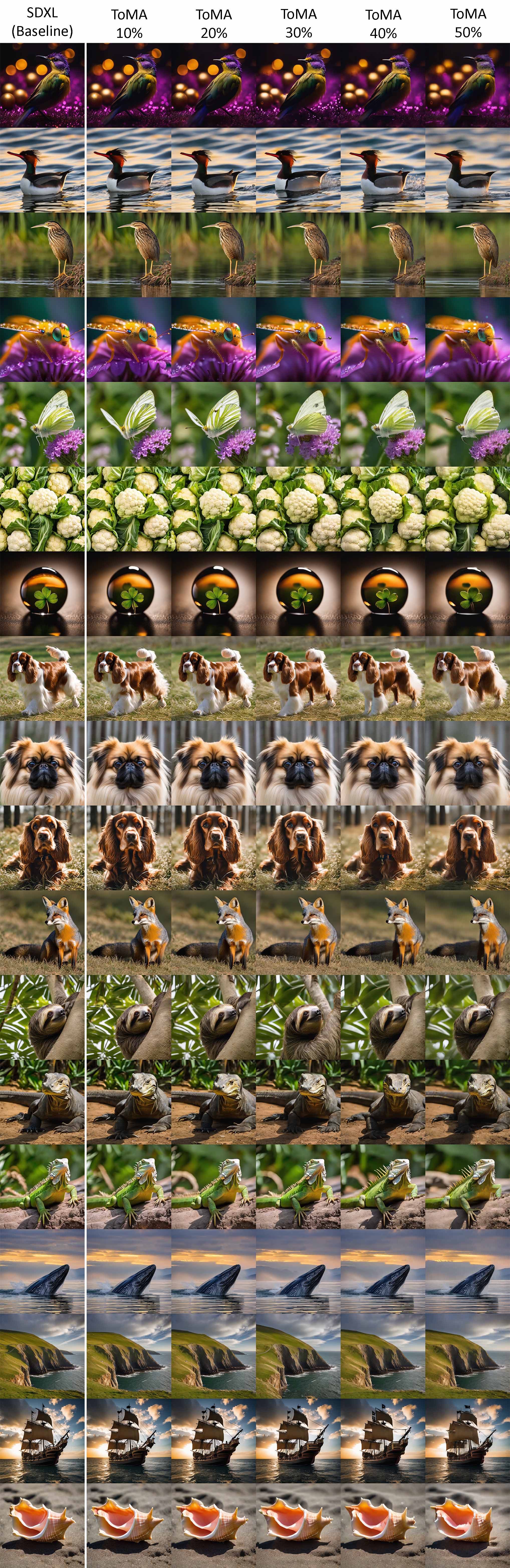}
		\label{fig:image1}
	\end{minipage}
	\hfill
	\begin{minipage}[t]{0.49\textwidth}
		\centering
		\includegraphics[width=0.85\textwidth]{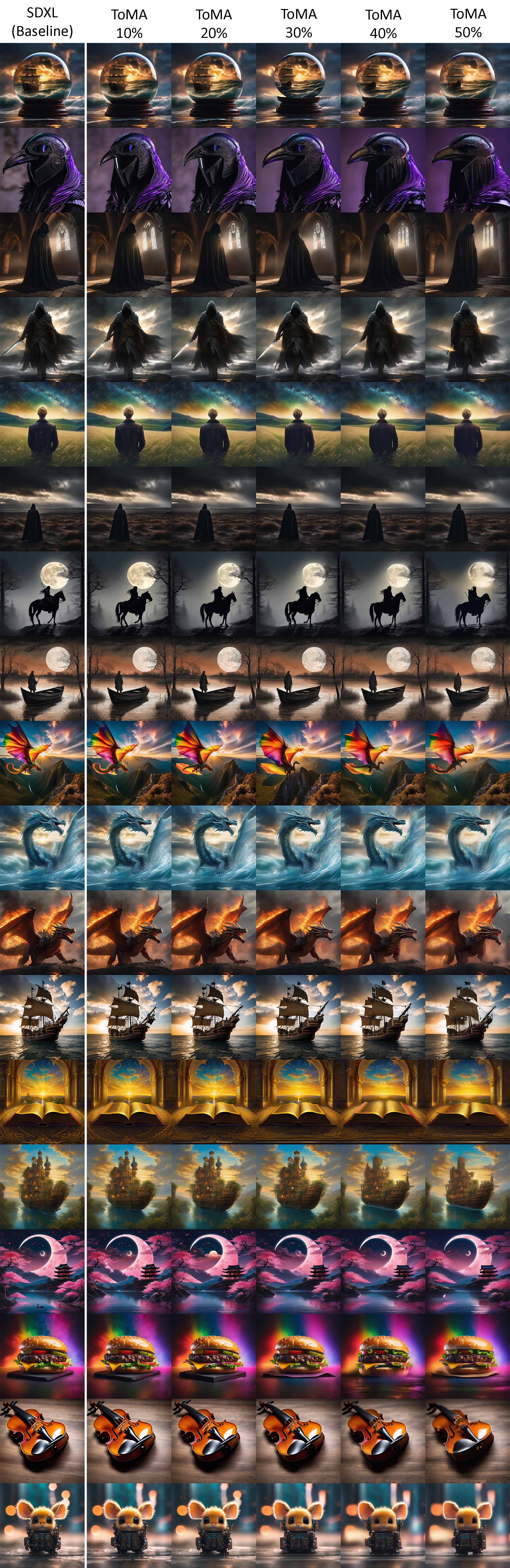}
		\label{fig:image2}
	\end{minipage}
	\vspace{-5pt}
	\caption{\small \textbf{Visual examples of ToMA}. Even with half of the tokens merged, ToMA consistently preserves image quality and often demonstrates greater robustness compared to other methods (ToDo, ToFu, and ToMeSD).}
	\label{fig: more on ToMA}
\end{figure}

We present additional visual comparisons between \texttt{SDXL}, ToMeSD, and ToMA below in Fig~\ref{fig: baseline models comparison}. The prompts are sampled from the GemRec and ImageNet-1K datasets. As the visuals reveal, ToMA maintains image quality more faithfully compared to other methods, especially in preserving fine-grained details and spatial coherence. This is evident in both synthetic scenes (e.g., fantasy landscapes, bowls of fire) and natural subjects (e.g., animals, boats, and portraits). ToMA Additional images generated with ToMA on \texttt{SDXL} are provided on the next page in Fig.~\ref{fig: more on ToMA}.  
\begin{figure}[htbp]
	\centering
	\includegraphics[width=0.8\textwidth]{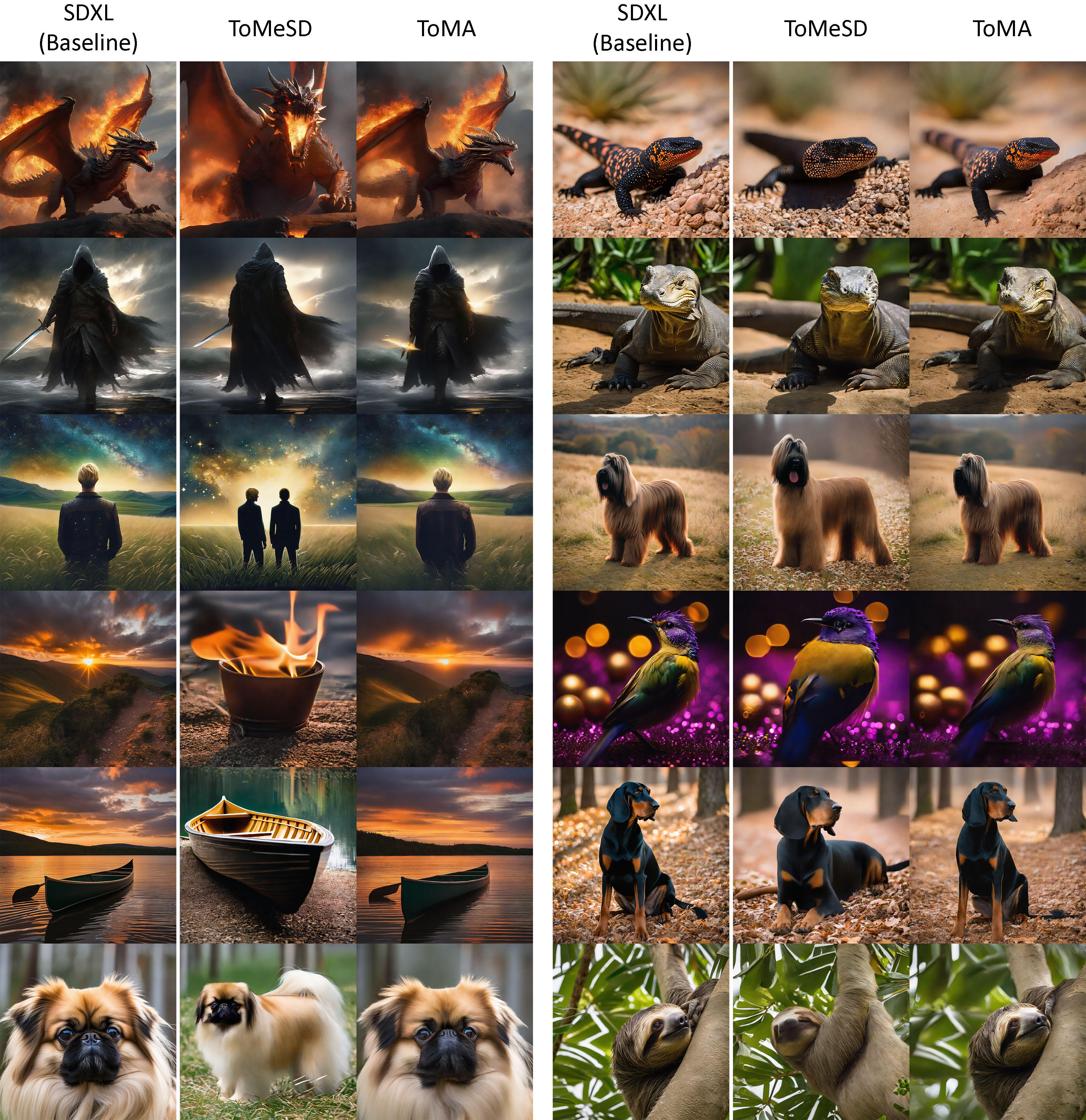}
	\caption{Qualitative comparison between Baseline \texttt{SDXL-base}, ToMeSD, and ToMA.}
	\label{fig: baseline models comparison}
\end{figure}

\vspace{-5pt}

\clearpage

\subsection{Images Generated with Flux}
\begin{figure}[htbp]
	\centering
	\includegraphics[width=0.8\textwidth]{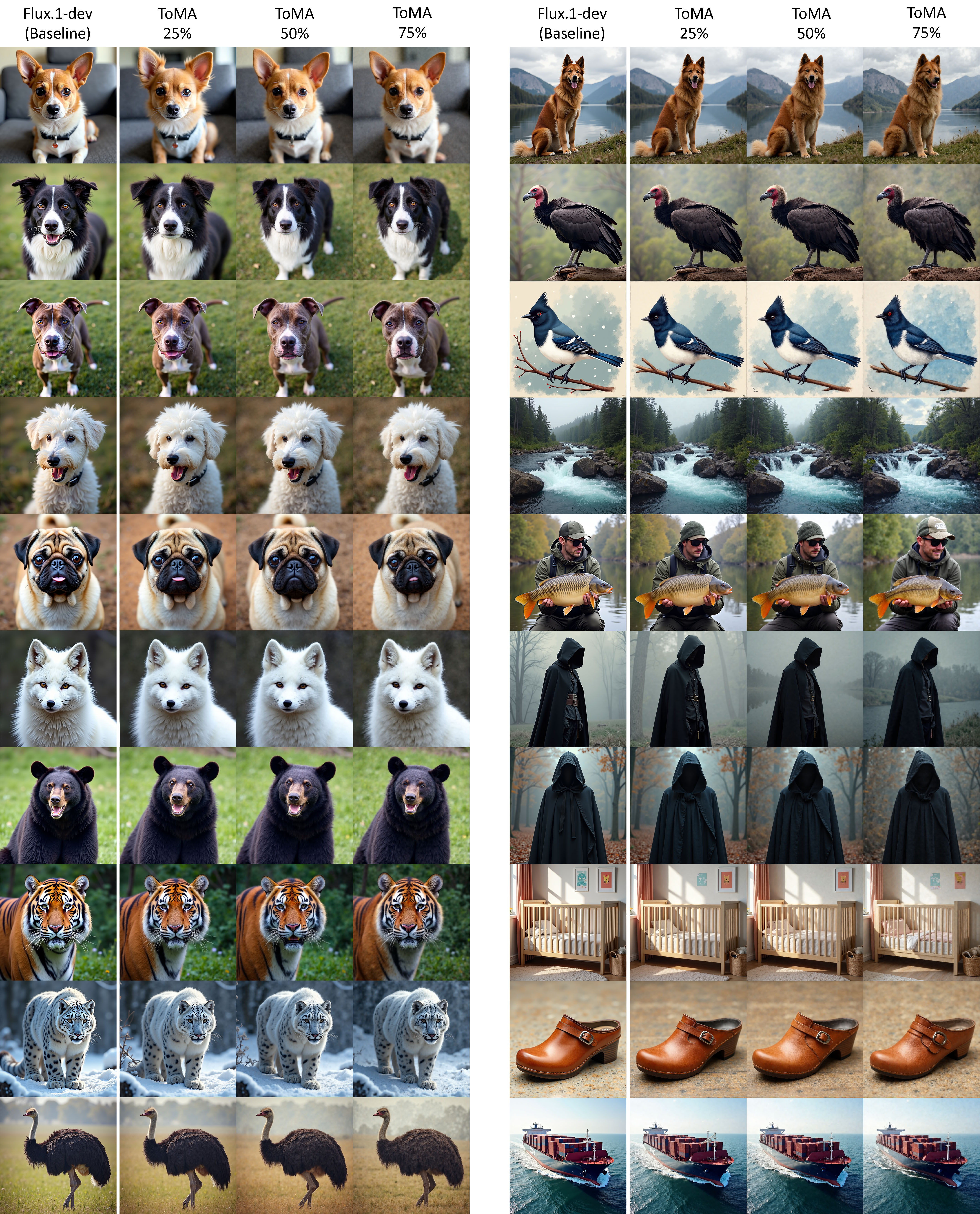}
	\caption{Qualitative comparison between Baseline \texttt{Flux.1-dev} and ToMA.}
	\label{fig: flux visual}
\end{figure}

Please refer to Fig.~\ref{fig: flux visual} below for more images generated with ToMA on \texttt{Flux1.0-dev}.

\clearpage
\section{Diffusion Transformers (DiT)}
\subsection{Locality in DiT}
\label{appendix: dit locality}

\begin{figure}[htbp]
  \centering
  \includegraphics[width=0.8\textwidth]{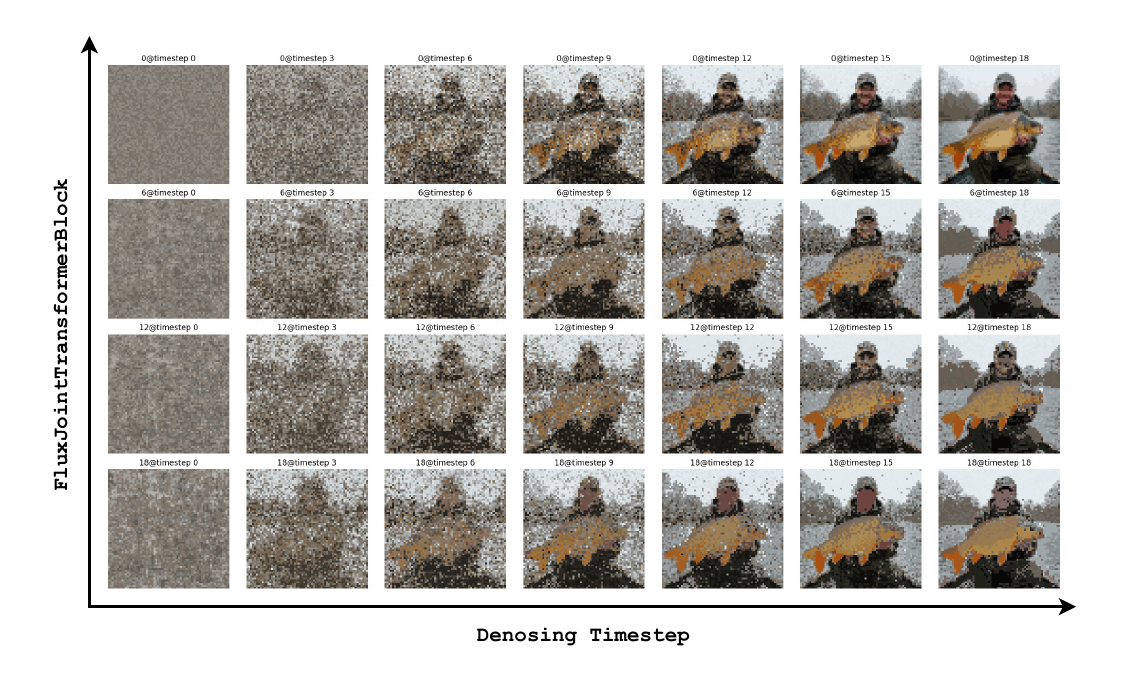}
  \caption{Re-colored K-means clusters of hidden states in \texttt{Flux.1-dev} across blocks and denoising steps.}
  \label{fig: dit locality}
\end{figure}

We inspected the hidden states of \texttt{Flux}.  
Using simple visualizations (K-means coloring) at the start of each block and across denoising timesteps, we observed that—even without convolutions—the hidden states already resemble the target image (Fig.~\ref{fig: dit locality}).  
This locality is introduced mainly by the rotary embeddings in \texttt{Flux}.
Empirically, when we apply submodular token selection \emph{within local windows}, the model still produces high-quality images.

\subsection{Transformers and Positional Embeddings in DiT}

DiT blocks differ from the usual “self-attention → cross-attention → MLP’’ pattern, so off-the-shelf merging methods such as ToMeSD, ToFu, or ToDo break the model (all-black / noisy and nonsense outputs).  
Two issues arise:

\begin{enumerate}
  \item The DiT block order (attention+MLP fusion) is not aligned with the assumptions in those methods.
  \item Positional embeddings (e.g.\ RoPE) are mixed with both image and text tokens; careless merging discards useful tokens.
\end{enumerate}

We therefore add two simple rules:

\begin{itemize}
  \item \textbf{Skip the first 10 blocks.}  
        Early blocks fuse text and image features; skipping them avoids over-merging.
  \item \textbf{Handle the two DiT block types separately.}
        \begin{description}
          \item[JointTransformer.]  
            Text and image tokens are projected \emph{separately}, then concatenated before RoPE.  
            We merge text and image tokens independently, then concatenate; RoPE indices are gathered accordingly.
          \item[SingleTransformer.]  
            Tokens are already concatenated.  
            We first split the hidden state back into text and image parts, merge each part, then re-concatenate; RoPE is gathered in the same way.
        \end{description}
\end{itemize}

These lightweight changes respect both modality boundaries and positional embeddings, allowing our token-merging variant to run on \texttt{Flux} with no perceptible loss in image quality while still delivering the intended speedup.

\newpage
\section{Ablation Study}
\label{Appendix: Ablation}

All the ablations reported in this appendix are conducted on \texttt{SDXL} with a default merge ratio of \(r=0.5\) unless otherwise noted.  Each experiment isolates one design choice so that its direct impact on quality—measured by CLIP, DINO, and pixel MSE, and on runtime—measured in seconds per generated image, all can be clearly assessed.

\subsection{Destination–selection strategy}
\label{sec:ablation:dst}

The first experiment evaluates four types of selection windows for choosing destination tokens: a global window that considers every pair of tokens, a local tile facility that performs the same selection inside non-overlapping windows, a horizontal stripe window, and a random baseline on the full window.  Tab.~\ref{tab:selection_methods} shows a clear pattern.  Restricting the search to tiles yields the best CLIP and DINO scores and the lowest MSE, while running more than six times faster than the exhaustive global search.  The result confirms our locality hypothesis from Fig.~\ref{fig: dit locality}: most informative tokens lie close to one another, so a global scan is unnecessary and wasteful.

\begin{table}[h]
  \centering
  \begin{tabular}{@{}r|cccc@{}}
    \toprule
    \textbf{Type} & \textbf{CLIP} $\uparrow$ & \textbf{DINO} $\downarrow$ & \textbf{MSE} $\downarrow$ & \textbf{Sec/img} $\downarrow$ \\
    \midrule
    Global          & 30.949 & 0.069 & 1{,}637 & 33.2 \\
    \cellcolor[gray]{0.9}Tile            & \cellcolor[gray]{0.9}31.019 & \cellcolor[gray]{0.9}0.055 & \cellcolor[gray]{0.9}1{,}274 & 5.1 \\
    Stripe          & 30.986 & 0.074 & 1{,}730 & 5.2 \\
    Random                   & 30.553 & 0.090 & 2{,}029 & \cellcolor[gray]{0.9}4.5 \\
    \bottomrule
  \end{tabular}
  \caption{Comparison of destination–selection rules.  Tile-based selection delivers the best quality and the lowest latency.  Values are averaged over the SDXL validation prompt set; CLIP/DINO rounded to three decimals, MSE to the nearest integer.}
  \label{tab:selection_methods}
\end{table}

\subsection{Tile granularity}

Once the tile–based facility was chosen, we next varied the number of tiles in order to control the spatial extent of each selection window.  Intuitively, fewer tiles correspond to larger windows and therefore allow tokens to compete across a wider context, whereas many small tiles enforce highly local competition.  Tab.~\ref{tab:tile_size_ablation} shows four granularities ranging from $4$ large tiles to $256$ small tiles.  Moving from $4$ to $16$ tiles yields a large quality jump: DINO improves by $17\%$ and MSE by $14\%$ while latency is cut almost in half.  The improvement continues when the window count rises to $64$, which records the best DINO and MSE and also the lowest runtime.  At $256$ tiles, DINO and MSE drift upward again, indicating that extremely small windows over-constrain the matching pool; nevertheless, CLIP remains tied with the best value, and the latency does not increase further because the GPU remains compute–bound.  

Because the numerical differences between $64$ and $256$ tiles are modest, $64$ tiles strike a cleaner balance: it delivers the strongest quality metrics while preserving the same throughput and avoiding the bookkeeping overhead that arises when thousands of tiny windows must be indexed.  For these reasons, we adopt $64$ tiles as the default granularity in the main paper.

\begin{table}[hbtp]
  \centering
  \begin{tabular}{@{}r|cccc@{}}
    \toprule
    \# \textbf{Tiles} & \textbf{CLIP} $\uparrow$ & \textbf{DINO} $\downarrow$ & \textbf{MSE} $\downarrow$ & \textbf{Sec/img} $\downarrow$ \\
    \midrule
     4   & 30.775 & 0.069 & 1{,}564 & 11.4 \\
    16   & 30.991 & 0.057 & 1{,}345 &  6.4 \\
    \rowcolor{gray!10}%
    64   & 31.019 & \cellcolor[gray]{0.9}0.055 & \cellcolor[gray]{0.9}1{,}274 & \cellcolor[gray]{0.9}5.0 \\
    256  & \cellcolor[gray]{0.9}31.027 & 0.057 & 1{,}296 & 5.0 \\
    \bottomrule
  \end{tabular}
  \caption{Influence of tile granularity at a 50\% merge ratio.  Using 64 tiles achieves the best DINO and MSE while matching the runtime of 256 tiles.}
  \label{tab:tile_size_ablation}
\end{table}

\subsection{Merge and unmerge latency}

The third experiment benchmarks the merge and unmerge kernels at a fixed sequence length of \(N = 1024\) tokens.  We compare the dense linear formulation used in ToMA with the index–scatter implementation in ToMeSD.  In ToMeSD the algorithm first builds a destination-index array, then gathers features with \texttt{torch.index\_select} and finally scatters them back with \texttt{index\_add\_}.  Because both gather and scatter operate on the full index list, their cost grows linearly with the merge ratio \(r\).  Moreover, the discontinuous memory accesses inherent in these calls leave many GPU warps idle.  

ToMA eliminates the two passes by replacing them with a single dense matrix multiplication \(\tilde{\mA}\mX\), where \(\tilde{\mA}\in\sR^{D\times N}\) and \(D = (1 - r)N\).  The operation therefore depends only on the output length \(D\); its cost is constant with respect to the number of removed tokens and maps efficiently to a single GEMM that fully utilizes GPU compute units.  As reported in Table~\ref{table:size_1024_comparison}, ToMA is consistently four to five times faster than ToMeSD for both merge and unmerge across all tested values of \(r\).

\begin{table}[hbtp]
  \centering
  \small
  \begin{tabular}{@{}l l|rrr|ccc@{}}
    \toprule
    \multirow{2}{*}{\textbf{Operation}} & \multirow{2}{*}{\textbf{Method}} &
      \multicolumn{3}{c|}{\textbf{Time} ($\mu$s)$\downarrow$} &
      \multicolumn{3}{c}{\textbf{Speedup}$\uparrow$} \\
    \cmidrule(lr){3-5}\cmidrule(lr){6-8}
    & & $25\%$ & $50\%$ & $75\%$ & $25\%$ & $50\%$ & $75\%$ \\
    \midrule\midrule
    \multirow{2}{*}{Merge} 
      & \texttt{ToMe} & 202.2 & 202.1 & 193.2 & -- & -- & -- \\ 
      & \texttt{ToMA}   & \cellcolor[gray]{0.9}39.0 & \cellcolor[gray]{0.9}38.8 & \cellcolor[gray]{0.9}38.8
               & \cellcolor[gray]{0.9}5.2$\times$ & \cellcolor[gray]{0.9}5.2$\times$ & \cellcolor[gray]{0.9}5.0$\times$ \\
    \midrule
    \multirow{2}{*}{Unmerge} 
      & \texttt{ToMe} & 160.5 & 160.1 & 144.0 & -- & -- & -- \\ 
      & \texttt{ToMA}   & \cellcolor[gray]{0.9}40.2 & \cellcolor[gray]{0.9}40.5 & \cellcolor[gray]{0.9}39.6
               & \cellcolor[gray]{0.9}4.0$\times$ & \cellcolor[gray]{0.9}3.9$\times$ & \cellcolor[gray]{0.9}3.6$\times$ \\
    \bottomrule
  \end{tabular}
  \caption{Micro-benchmarks at sequence length 1024 (median over 1{,}000 runs on an NVIDIA RTX6000) across different merge ratios.  Shaded cells indicate the best result per column.  ToMA is roughly four to five times faster than ToMeSD over all merge ratios \(r\).}
  \label{table:size_1024_comparison}
\end{table}

\subsection{Transpose \textit{versus} pseudo--inverse for unmerge}

We experiment whether a mathematically exact unmerge, obtained via the Moore--Penrose pseudo--inverse of the merge matrix, offers any quality advantage over the much cheaper transpose.  
In theory, the pseudo--inverse should restore the pre–merge feature space more faithfully, because it inverts the least–squares projection implicit in the merge.  
In practice, the merge matrix used by ToMA is highly sparse and close to orthogonal, so the transpose already provides an excellent approximation.  
Computing the pseudo--inverse requires a QR or SVD decomposition of the \(D\times N\) merge matrix, followed by two matrix multiplications to apply the result.  
These decompositions are memory--bandwidth bound and cannot be fused with the surrounding transformer layers, so the cost is paid in every unmerge step.

Table~\ref{table:unmerge_comparison} confirms that the extra work is wasted.  
Across 300 generated images, the pseudo--inverse gains no measurable improvement: CLIP, DINO, and MSE differ by less than \(1\%\).  
Meanwhile, latency more than doubles because the decomposition incurs additional global synchronizations on the GPU.  
Given the negligible benefit and the clear timing penalty, we adopt the simple transpose as the default unmerge method.

\begin{table}[hbtp]
  \centering
  \begin{tabular}{@{}l|cccc@{}}
    \toprule
    \textbf{Unmerge Method }& \textbf{CLIP} $\uparrow$ & \textbf{DINO} $\downarrow$ & \textbf{MSE} $\downarrow$ & \textbf{Sec/img} $\downarrow$ \\
    \midrule
    \rowcolor{gray!10}%
    Transpose        & 31.027 & 0.057 & 1{,}296 & 4.8 \\
    Pseudo--inverse   & 30.997 & 0.057 & 1{,}288 & 10.1 \\
    \bottomrule
  \end{tabular}
  \caption{Transpose versus pseudo--inverse unmerge at 50\% merge.  Quality metrics are identical, but transpose is more than twice as fast.}
  \label{table:unmerge_comparison}
\end{table}

\subsection{Recompute schedule}

The final ablation varies how often destination indices and attention weights are recomputed during denoising.  Tab.~\ref{table:recompute_schedule} indicates that refreshing attention every step gives the best overall accuracy, whereas destination indices can be updated ten times less frequently with minimal loss.  A schedule of “destination every 10 steps, attention every 5 steps’’ preserves 99\% of the peak quality while roughly halving recompute FLOPs, and is therefore adopted in the main experiments.

\begin{table}[hbtp]
  \centering
  \small
  \begin{tabular}{@{}l|l|cccc@{}}
    \toprule
    \textbf{Recompute} \(\mD\) & \textbf{Recompute} \(\tilde{\mA}\) &
    \textbf{CLIP} $\uparrow$ & \textbf{DINO} $\downarrow$ & \textbf{MSE} $\downarrow$ & \textbf{Sec/img} $\downarrow$ \\
    \midrule
    Every 50 steps & Every 50 steps & 30.043 & 0.077 & 2{,}489 & \cellcolor{gray!15}4.84 \\
    Every 10 steps & Every 10 steps & 30.817 & 0.073 & 1{,}735 & 4.97 \\
    \rowcolor{gray!8}
    Every 10 steps & Every \phantom{0}5 steps & 30.865 & 0.070 & 1{,}632 & 5.00 \\
    Every 10 steps & Every \phantom{0}1 step  & \cellcolor{gray!15}30.997 & \cellcolor{gray!15}0.067 & \cellcolor{gray!15}1{,}525 & 5.06 \\
    Every \phantom{0}5 steps & Every \phantom{0}5 steps & 30.892 & 0.069 & 1{,}609 & 4.92 \\
    Every \phantom{0}1 step  & Every \phantom{0}1 step  & 30.920 & 0.067 & 1{,}552 & 5.05 \\
    \bottomrule
  \end{tabular}
  \caption{Effect of recomputation frequency at 50\% merge.
  The shaded cells denote the best metric in each column.}
  \label{table:recompute_schedule}
\end{table}

\newpage
\section{Memory analysis}
\label{sec: memory}
Table~\ref{tab:max_memory} provides a peak–memory audit on our methods.  
For each model, we record both the maximum allocated memory, which reflects the live tensor footprint, and the maximum reserved memory, which includes CUDA’s internal caching.  
Across all three merge ratios the numbers remain tightly clustered: on \texttt{Flux}, the largest deviation from the dense baseline is a 0.3\% increase in allocated memory for plain ToMA at 25\% merging; on \texttt{SDXL-base}, the worst case is a 1.9\% rise in reserved memory, again for plain ToMA at 25\%.  
The tile variant is even closer and occasionally dips \emph{below} the baseline because smaller activation tensors leave more room for the allocator to reuse blocks.  

\begin{table}[h]
  \centering
  \small

\begin{tabular}{@{}c|c|l|ccc@{}}
  \toprule
  \multirow{2}{*}{\textbf{Model}} 
    & \multirow{2}{*}{\textbf{Metric}} 
    & \multirow{2}{*}{\textbf{Method}} 
    & \multicolumn{3}{c}{\textbf{Max Memory (MB)}$\downarrow$} \\
  \cmidrule{4-6}
    &   &   & 25\% & 50\% & 75\% \\
  \midrule
  \multirow{6}{*}{\texttt{Flux.1-dev}}
    & \multirow{3}{*}{Alloc.} 
        & \texttt{Baseline}           & \cellcolor[gray]{0.9}34,640 & \cellcolor[gray]{0.9}34,640 & \cellcolor[gray]{0.9}34,640 \\
    &   & \texttt{ToMA}               & 34,744 & 34,710 & 34,675 \\
    &   & \texttt{ToMA\textsubscript{tile}}  & 34,647 & 34,647 & 34,642 \\
  \cmidrule{2-6}
    & \multirow{3}{*}{Resv.} 
        & \texttt{Baseline}           & \cellcolor[gray]{0.9}37,002 & \cellcolor[gray]{0.9}37,002 & \cellcolor[gray]{0.9}37,002 \\
    &   & \texttt{ToMA}               & 37,050 & 36,976 & 36,954 \\
    &   & \texttt{ToMA\textsubscript{tile}}  & 37,054 & 37,006 & 36,950 \\
  \midrule
  \midrule
  \multirow{8}{*}{\texttt{SDXL-base}}
    & \multirow{4}{*}{Alloc.}
        & \texttt{Baseline}           & \cellcolor[gray]{0.9}10,721 & \cellcolor[gray]{0.9}10,721 & \cellcolor[gray]{0.9}10,721 \\
    &   & \texttt{ToMA}               & 10,931 & 10,857 & 10,797 \\
    &   & \texttt{ToMA\textsubscript{stripe}}  & 10,722 & 10,719 & \cellcolor[gray]{0.9}10,718 \\
    &   & \texttt{ToMA\textsubscript{tile}}    & 10,725 & \cellcolor[gray]{0.9}10,720 & 10,719 \\
  \cmidrule{2-6}
    & \multirow{4}{*}{Resv.}
        & \texttt{Baseline}           & 14,150 & 14,150 & 14,150 \\ 
    &   & \texttt{ToMA}               & 14,460 & 14,260 & \cellcolor[gray]{0.9}14,130 \\
    &   & \texttt{ToMA\textsubscript{stripe}}  & \cellcolor[gray]{0.9}14,114 & 14,188 & 14,222 \\
    &   & \texttt{ToMA\textsubscript{tile}}    & 14,158 & \cellcolor[gray]{0.9}14,158 & 14,182 \\
  \bottomrule
\end{tabular}
  \caption{Peak GPU memory (MB) for different \texttt{ToMA} variants on \texttt{Flux.1-dev} and \texttt{SDXL-base} across three merge ratios. We report both maximum allocated and reserved memory. ($\downarrow$: lower is better.)}
  \label{tab:max_memory}
\end{table}

\section{FLOP Analysis}
\label{sec: flop}

Table~\ref{tab:flop-singlecol} reports a layer-wise floating-point-operation (FLOP) breakdown, restricted to the dominant computational modules inside each transformer block: the $QKV$\,/\, output projections and the attention matrix products.  Results are shown for the largest block in \texttt{Flux} and for the two block types that occur in \texttt{SDXL}.  Applying ToMA with a 50\% merge ratio yields a $2.3\times$ reduction on \texttt{Flux} and up to a $3.4\times$ reduction on \texttt{SDXL}.  The additional FLOPs introduced by ToMA—namely submodular token selection, the merge weight computation, and the linear (un)merge kernels—amount to less than~1\,\% of the new total and are therefore negligible at the scale of the overall savings.

\begin{table}[h]
  \centering
  \small
  \begin{tabular}{@{}l|c|ccc|c@{}}
    \toprule
    \multirow{2}{*}{\textbf{Model}} & \textbf{Layer Size} & \multicolumn{3}{c|}{\textbf{FLOPs (G)} $\downarrow$} &  \multirow{2}{*}{\textbf{Reduction}} \\
    \cmidrule(lr){3-5}
                   & (Seq\,$\times$\,Dim) & \textbf{Original} & \textbf{ToMA (50\%)} & \textbf{Overhead} &  \\
    \midrule
    \texttt{Flux}  & 4608\,$\times$\,3072 & 520 & 225 & 1.01 & $\sim$2.3$\times$ \\
    \texttt{SDXL}  & 4096\,$\times$\,640\phantom{0}  & 106 &  32 & 0.42 & $\sim$3.4$\times$ \\
    \texttt{SDXL}  & 1024\,$\times$\,1280 &  30 &  13 & 0.06 & $\sim$2.4$\times$ \\
    \bottomrule
  \end{tabular}
  \caption{Layer-level FLOP counts before and after applying ToMA at a 50\% merge ratio.  The “Overhead’’ column includes sub-modular selection, merge weight computation, and the linear (un)merge operations.}
  \label{tab:flop-singlecol}
\end{table}


\end{document}